%% file: main.tex
\documentclass{article}

\PassOptionsToPackage{table}{xcolor}
\PassOptionsToPackage{numbers,sort&compress}{natbib}
\usepackage[preprint]{neurips_2026}

\usepackage[utf8]{inputenc}
\usepackage[T1]{fontenc}
\usepackage{graphicx}
\usepackage{algorithm}
\usepackage{algorithmic}
\usepackage{subcaption}
\usepackage[most]{tcolorbox}
\usepackage{enumitem}
\usepackage{multirow}
\usepackage{booktabs}
\usepackage{tabularx}
\usepackage{amsmath}
\usepackage{amssymb}
\usepackage{mathtools}
\usepackage{amsthm}
\usepackage{float}
\usepackage[hidelinks]{hyperref}
\usepackage[capitalize,noabbrev]{cleveref}

\graphicspath{{./}{./figures/}}

\theoremstyle{plain}

\theoremstyle{definition}

\theoremstyle{remark}

\title{A Neuropsychologically Grounded Evaluation of LLM Cognitive Abilities}

\author{%
  Faiz Ghifari Haznitrama \quad Faeyza Rishad Ardi \quad Alice Oh \\
  School of Computing, KAIST, Daejeon, South Korea \\
  \texttt{haznitrama@kaist.ac.kr}
}

\begin{document}

\maketitle

\begin{abstract}
  \input{sections/0_abstract}
\end{abstract}

\section{Introduction}
\label{sec:introduction}
\input{sections/1_introduction}

\section{Related Works}
\label{sec:related_works}
\input{sections/2_related_works}

\section{NeuroCognition Benchmark Tests}
\label{sec:neurocognition}

\subsection{Raven's Progressive Matrices}
\label{sec:rapm}
\input{sections/3_1_rapm}

\subsection{Spatial Working Memory}
\label{sec:swm}
\input{sections/3_2_swm}

\subsection{Wisconsin Card Sorting Test}
\label{sec:wcst}
\input{sections/3_3_wcst}

\subsection{Experiment Setup}
\label{sec:experiment-setup}
\input{sections/3_4_experiment_setup}

\section{Results}
\label{sec:results}

% Table~\ref{tab:main-results} reports the baseline results for NeuroCognition, with complete results in Appendix \ref{sec:full-results}.

\subsection{Key Findings}
\label{sec:key-findings}
\input{sections/4_1_key_findings}

\subsection{Failure Cases \& Patterns}
\label{sec:failure-cases-patterns}
\input{sections/4_2_failure_cases_patterns}

\section{Ablation Study}
\label{sec:ablation}

% Following the baseline results, we explore aspects that could affect model performance. We present the change in performance compared to Table \ref{tab:main-results} baseline using a subscripted delta ($\Delta$) notation to indicate any improvements or declines.

\subsection{RAPM: Reasoning vs No Reasoning}
\label{sec:reasoning-effect-rapm}
\input{sections/5_1_effect_reasoning_rapm}

\subsection{RAPM: Pattern Hints}
\label{sec:improvement-rapm}
\input{sections/5_2_improve_pattern_rapm}

\subsection{SWM \& WCST: Note-Taking Assistance}
\label{sec:improvement-swm-wcst}
\input{sections/5_3_improve_notes_swm_wcst}

\section{Analysis \& Discussion}
\label{sec:analysis-discussion}

\subsection{Factor Analysis of LLM}
\label{sec:factor-analysis-llm}

\input{sections/6_1_factor_analysis_llm}
\begin{figure}[!htbp]
  \centering
  \small
  \includegraphics[width=\linewidth]{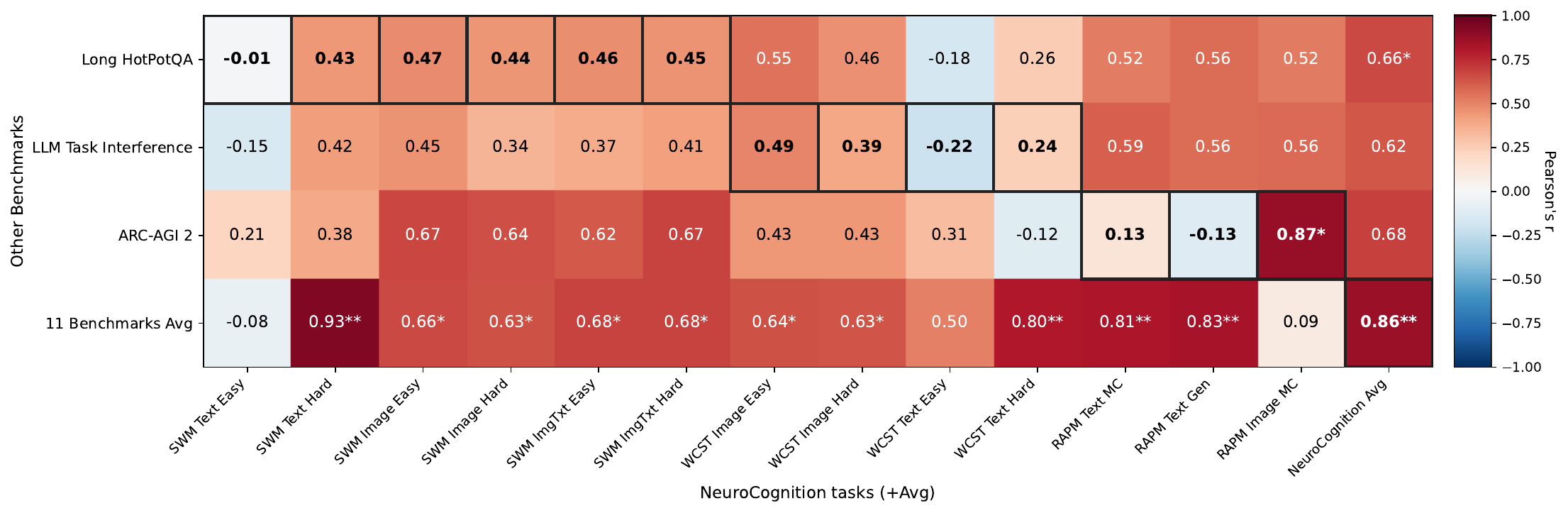}
  \caption{Correlations between NeuroCognition tests and real-world LLM benchmarks. Bold cells with highlighted borders indicate task pairs relying on the same cognitive abilities. Asterisks denote significance ($*p < .05$, $**p < .01$).}
  \label{fig:corr-all}
\end{figure}
\subsection{Correlation with Other Benchmarks}
\label{sec:correlation-other-benchmarks}
\input{sections/6_2_correlation_other_benchmarks}

\subsection{Human-LLM Alignment}
\label{sec:construct-validity}
\input{sections/6_3_relation_with_human}

\section{Conclusion}
\label{sec:conclusion}
\input{sections/7_conclusion}

% TODO: revert heading to default ("Acknowledgments and Disclosure of Funding") once funding info is added
\renewcommand{\acksection}{\section*{Acknowledgments}}
\begin{ack}
We thank Afrizal Hasbi Azizy for the insightful consultation and feedback from cognitive science perspectives. We also thank the participants for their willingness to participate in this study.
\end{ack}

\bibliographystyle{plainnat}
\bibliography{custom}

\raggedbottom
\appendix

\input{appendix/human_protocol}
\input{appendix/human_results}
\input{appendix/rapm_text_details}
\input{appendix/experiment_setup}
\input{appendix/benchmark_setup}

\input{appendix/factor_analysis}

\input{appendix/full_results}
\input{appendix/prompts}
\input{appendix/examples}

% Ablation figure now in main paper (Section~\ref{sec:ablation}).
% \input{appendix/ablation_figures}
% \input{appendix/ablation_tables}

\clearpage
\section{Limitations}
\label{sec:limitations}
\input{sections/limitations}

\section{Broader Impact}
\label{sec:impact-statement}
\input{sections/impact_statement}

\section{Future Work}
\label{sec:future-work}
\input{sections/8_future_work}

\end{document}

%% file: sections/0_abstract.tex
Large language models (LLMs) display a unified "general factor" of capability across 10 benchmarks (a finding confirmed by our factor analysis of 156 models), yet they still struggle with simple, trivial tasks for humans. This is because current benchmarks focus on task completion, failing to probe the foundational cognitive abilities that highlight these behaviors. We address this by introducing the NeuroCognition benchmark, grounded in three adapted neuropsychological tests targeting \textbf{distinct foundational cognitive components}: Raven’s Progressive Matrices (abstract relational reasoning), Spatial Working Memory (goal-directed spatial updating), and the Wisconsin Card Sorting Test (cognitive flexibility). Our evaluation reveals that while models perform strongly on text, their performance degrades for images and with increased complexity. Comparison with a human baseline shows that LLMs and humans fail at different parts of the same tasks.
% Comparison with a human baseline shows humans follow the neuropsychologically predicted ordering while LLMs invert it, and humans and LLMs fail at different parts of the same tasks.
%  a construct-level inversion, where
Furthermore, we observe that complex reasoning is not universally beneficial, whereas simple, human-like strategies yield partial gains. We also find that NeuroCognition correlates positively with standard general-capability benchmarks, while still measuring distinct cognitive abilities beyond them. Overall, NeuroCognition emphasizes where current LLMs align with human-like intelligence and where they lack core adaptive cognition, showing the potential to serve as a verifiable, scalable source for improving LLMs.

%% file: sections/1_introduction.tex
Large language models (LLMs) show increasingly powerful capabilities \citep{achiam2023gpt}, further enhanced by the emergence of reasoning abilities \citep{wei2022chainofthought}. This is demonstrated by their beyond human-level capabilities in tasks such as olympiad-level problems \citep{trinh2024solving}. However, even with these impressive capabilities, they still stumble on trivial or straightforward tasks \citep{srivastava2022beyond, lewis2024usingcounterfactualtasksevaluate}. This raises a curiosity about the underlying cognitive mechanism of LLMs and how it compares to human cognitive abilities.

Most current LLM benchmarks are task-specific, limiting their analysis and insights to the task performance itself. These benchmarks are not able to uncover the underlying cognitive abilities required to solve the task \citep{lin2024criticbench, honeyhive2024pitfalls} because task performance is not a reliable measure of intelligence, but rather heavily influenced by prior knowledge and experience \citep{chollet2019measureintelligence}.
%There is also a lack of benchmarks that focus on aspects where LLMs are still struggling, such as long multi-turn conversations \citep{laban2025llmslostmultiturnconversation, kwan2024mtevalmultiturncapabilitiesevaluation, sirdeshmukh2025multichallengerealisticmultiturnconversation}, sustained reasoning \citep{patel2024multilogievalevaluatingmultisteplogical, lei2025mscorebenchmarkmultistagecollaborative}, and multimodal understanding \citep{zheng2023judging, ge2024mllmbenchevaluatingmultimodalllms}.
One interesting test that is often overlooked is neuropsychological tests, which are a well-established method for measuring specific cognitive abilities in humans, originally designed to isolate specific cognitive functions using controlled stimuli and interpretable outcomes. Previous works have attempted to apply neuropsychological tests to LLMs \citep{gong2024workingmemorycapacitychatgpt, hao2025visuallargelanguagemodels}, but they are limited to a single task or modality, or they lack a deeper analysis of the model's underlying cognitive abilities.

\begin{figure*}[t]
  \centering
  \includegraphics[width=\linewidth]{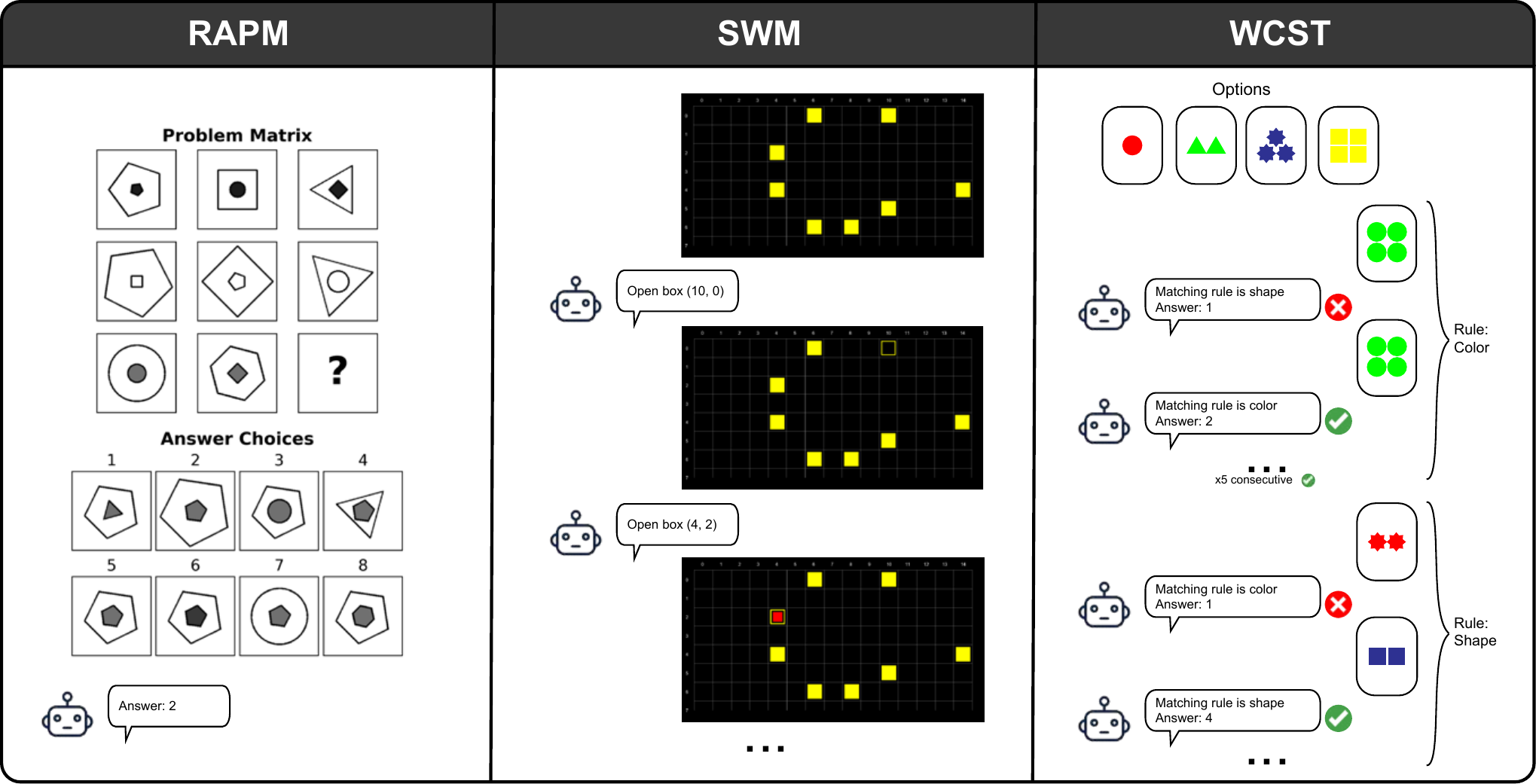}
  \caption{The three neuropsychological tests used in our work.}
  \label{fig:neuro-cognition-task}
\end{figure*}

We present the NeuroCognition benchmark, built by adapting from three well-established neuropsychological tests targeting distinct foundational cognitive components: Raven's Progressive Matrices \citep{raven1938raven, raven1938progressive, Raven1941-hg, carpenterrapm} (RAPM, fluid intelligence), Spatial Working Memory \citep{baddeley1974working} (SWM, goal-directed spatial updating), and the Wisconsin Card Sorting Test \citep{Berg1948-hj, grant1993wisconsin} (WCST, cognitive flexibility). We adopt these tests as a probe of distinct foundational components in LLMs. However, adapting these tests for LLMs requires some adjustments, specifically for SWM and WCST, because LLMs are stateless and need to be provided with the history of past moves, similar to a process known as \textit{cognitive offloading} \citep{risko2016cognitive, larkin1980expert}.
% Unlike humans who keep track of task state in working memory between steps, LLMs by default do not have a persistent memory between turns, so we provide in-context history of past moves.
Therefore, rather than memorizing, the model then reasons over the provided history to track state of the test implicitly. This makes the tests not fully equivalent to human, but the core cognitive demands still apply as the tests still target the same underlying cognitive abilities, just without the memory component that replaced by the long in-context history. Each test is further adapted into multiple text and image variants (including harder versions) to vary modality and complexity, and we include experiments with human-like assistance strategies (e.g., pattern hints and notes) to assess whether they reduce errors and improve performance.

Our main contributions are:
\begin{itemize}[itemsep=0.1em, topsep=0.1em]
  \item NeuroCognition, a benchmark repurposed from well-established neuropsychological tests, into a simple, multimodal, scalable, robust assessment, and easily reproducible suite with various setups.
  \item Evaluation results show that state-of-the-art models perform strongly in baseline text settings, but struggle in image or when task complexity increases, and that explicit reasoning is not always beneficial.
  \item Human baseline comparison ($N{=}11$) reveals humans and LLMs fail at different parts of the same tasks, and only the top two models that show a sign of human-like cognitive sensitivity.
  \item Experiments with human-like strategies (pattern hints and notes) show mixed effects, indicating that the core issue lies in the models' fundamental cognitive abilities.
  \item Factor analysis and correlation results show that NeuroCognition measures specific cognitive abilities that differ from, yet correlate with, the general capabilities captured by standard benchmarks.
\end{itemize}

Overall, NeuroCognition is a benchmark that can identify gaps in LLMs' cognitive abilities relative to humans, offering insights into how LLMs operate and informing the design and training of future LLMs. Our code, data, and results are publicly available at \url{https://github.com/reggans/NeuroCognition}.

%% file: sections/2_related_works.tex
The neuropsychological tests included in our work measure different aspects of human cognitive abilities. The RPM \citep{raven1938raven, raven1938progressive, Raven1941-hg, carpenterrapm} measures abstract, relational reasoning capabilities \citep{snow1984topography}. RPM has been used to identify intellectually gifted individuals, and has been shown to be widely reliable across different geographical boundaries \citep{Mills01021993}. The SWM test \citep{baddeley1974working}, on the other hand, measures the working memory, the ability to hold and mentally work with information, which may no longer be perceptually present \citep{Diamond2013}. This test has been used to detect at-risk individuals for psychosis \citep{WOOD_PANTELIS_PROFFITT_PHILLIPS_STUART_BUCHANAN_MAHONY_BREWER_SMITH_McGORRY_2003}, and highlight the working memory deficits in individuals with autism \citep{steele2007spatial}. 
In the original CANTAB SWM, participants must internally maintain the state of the test. 
% requires a decision about whether to provide this history in context. 
Adapting SWM to LLMs by providing historical context removes the maintenance demand and isolates the manipulation demand of working memory, which we use here to mean selecting and updating information and acting on it toward a goal \citep{baddeley1974working}, and which we describe further in Section~\ref{sec:swm}. Finally, WCST \citep{Berg1948-hj, grant1993wisconsin} measures cognitive flexibility, i.e., the ability to adapt to a change in perspective and reallocate resources to process relevant information \citep{Diamond2013, Giller2019}. Patients with schizophrenia are shown to perform worse on this test \citep{everett2001performance}, and are shown to be robust even in clinical settings \citep{doi:10.1177/1073191119866257}.

Recent advances in LLMs have led many works seeking to establish stronger correlations between LLMs and human intelligence. Many works focus on developing a cognitive benchmark aiming to measure fluid intelligence \citep{chollet2019measureintelligence, gendron2024largelanguagemodelsstrong, yang2025trulyassessingfluidintelligence, pandya2025ntsebenchcognitivereasoningbenchmark}, emphasizing abstractness and dynamic evaluation to avoid the data contamination issue \citep{li2024latestevaladdressingdatacontamination}. Other works applied human cognitive tests to LLMs \citep{ dasgupta2024languagemodelshumanlikecontent} and showcased cognitive behaviors in LLMs, such as similar working memory capacity between humans and GPT-4 in the n-back test \citep{gong2024workingmemorycapacitychatgpt}, and above human-level cognitive flexibility in the WCST \citep{hao2025visuallargelanguagemodels}. CogBench \citep{codaforno2024cogbench} benchmarks 40 LLMs across seven cognitive psychology experiments using ten behavioral metrics, finding that model size and RLHF are the main drivers of human-aligned behavior. 
% Unlike CogBench, NeuroCognition uses clinical-grade neuropsychological tests with multimodal variants and process-aware metrics that link failure patterns to specific cognitive components.
 Some works also revealed common factors of models' performance across different benchmarks \citep{burnell2023revealingstructurelanguagemodel, ILIC2024101858}, indicating the existence of a common underlying intelligence.

Aside from cognitive abilities, several studies have also quantified specific areas where LLMs struggle. In multi-turn evaluation, LLMs tend to perform worse than single-turn evaluations \citep{laban2025llmslostmultiturnconversation}, and still perform sub-optimally in multi-turn tasks \citep{kwan2024mtevalmultiturncapabilitiesevaluation, sirdeshmukh2025multichallengerealisticmultiturnconversation}. LLMs also struggle in longer multi-step reasoning \citep{patel2024multilogievalevaluatingmultisteplogical} and in complex multi-stage tasks \citep{lei2025mscorebenchmarkmultistagecollaborative}. LLMs also exhibit cross-modal hallucination \citep{villa2025magicmerlimmultimodalevaluation}, and struggle in multi-image, multi-turn benchmarks \citep{liu2024mmdu, yan2025mmcradvancingvisuallanguage}. LLMs also fail in simple pattern tasks using modified alphabets \citep{lewis2024usingcounterfactualtasksevaluate}, indicating the lack of general analogical reasoning in LLMs.

%% file: sections/3_1_rapm.tex
\textbf{Task.} RPM presents a 3x3 matrix of construct patterns with the bottom-right cell missing. The participant must infer abstract relations across rows and columns, then fill the missing cell. 
% by either selecting one of the 8 available options or constructing it.

\textbf{Setup.} For the image version, we use the RAVEN dataset \citep{zhang2019raven}, implementation of RAPM with 7 different figure configurations licensed with GPL-3.0. We take 20 items per configuration, resulting in a total of 140 test data. We combine the 3x3 question matrix with the 8 available options into a single image input, as shown in Figure \ref{fig:neuro-cognition-task}, and present it to the model in multiple-choice (MC) format. For the text setting, we programmatically generate symbolic matrices using character and string attributes (see Appendix \ref{appendix:text-rapm-generation} for generation details). We provide two answer formats for text setting: multiple-choice (MC, same 8 options) and generative (Gen, free-form string). A stored rule representation enables automatic validation of Gen outputs (see Appendix \ref{appendix:text-rapm-validation} for validation details). We generate 200 test data for the text setting with random configurations.

\textbf{Construct Adaptation.} RPM is the canonical measure of \textit{fluid intelligence} (Gf), which defined as perceiving relations, reasoning abstractly, and solving novel problems independent of prior knowledge \citep{cattell1943measurement,cattell1963theory,carpenterrapm}. Since Gf is modality-independent by definition, we can argue that both our image and text variants of RAPM measure the same construct.

\textbf{Metrics.} We report accuracy for both MC and Gen. Average human performance for images is 0.814, reported by \citet{zhang2019raven}. For the text variant, a pilot ($N{=}9$ each) shows mean accuracy $0.430_{\pm.167}$ (MC) and $0.444_{\pm.183}$ (Gen), exceeding chance ($1/8$) and indicating the items are solvable but non-trivial (Appendix~\ref{appendix:human-results}).

%% file: sections/3_2_swm.tex
\textbf{Task.} In the SWM test, participants are presented with an image with several boxes. A token is hidden in one of the boxes, and participants are asked to search for it by elimination, opening one box at a time. After the token is found, it is regenerated in another box that has not previously contained it in the current trial. This is repeated until all boxes have contained all tokens, with the goal of locating all tokens with minimal redundant checking and without revisiting boxes unnecessarily.

\textbf{Setup.} We implement two difficulty levels. \textbf{Easy}: 8 boxes, 1 token active at a time, early stop after all 8 tokens found or a maximum of 64-guess cap. \textbf{Hard}: 12 boxes with 2 distinct tokens active concurrently (24 tokens found required, maximum 144 guesses), with each token directly reappearing after being found, and one box can contain both tokens at once. We experiment with three types of feedback setups: text-only, image-only (a grid with no additional textual information), and image+text variants. The image input uses a grid system, as shown in Figure \ref{fig:neuro-cognition-task}. The text-only setup identifies the boxes by number and ask the model to choose a box number.
%  for easier answer verification

\textbf{Construct adaptation.} The original CANTAB SWM \citep{Diamond2013} probes working memory \citep{baddeley1974working}, which has two demands. The first is \textit{maintenance}, holding information in a limited-capacity buffer against interference. The second is \textit{manipulation}, which we use here to mean selecting and updating information and acting on it toward a goal. Since LLMs are stateless, we provide move history across turns in context, which substitutes (1) with long-context reading and isolates (2). Therefore, our adapted SWM still targets the same underlying ability, even though the protocol is slightly different. An alternative is to update and pass an explicit state summary each turn . However, it provides the model with a processed summary of the current state, while in-context history requires the model to infer that state manually, which ensures that information manipulation remains the primary challenge in our evaluation.
% explicit state tracking pre-digests the current state for the model, while in-context history preserves the need to infer it --- keeping manipulation as the active bottleneck in our evaluation.
% Our adapted SWM therefore still targets the same underlying ability of goal-directed spatial updating, even though the surface protocol differs.

\textbf{Metrics.} Since SWM is highly affected by luck, we define some metrics that focus on penalizing errors and unfinished tests. We classify three avoidable error types:
\begin{itemize}[itemsep=0.1em, topsep=0.1em]
  \item \textbf{illegal}: opens a box that can no longer contain a token because it has already yielded all available tokens for the trial.
  \item \textbf{no-box}: refers to a box index/identifier that does not exist.
  \item \textbf{repeated}: re-opens a box that was previously opened and found empty, before any new token was regenerated (i.e., no state change that could justify revisiting).
\end{itemize}
These three error types sum to $n_{err}$. Let $n_{valid}$ be the total valid guesses. A consistency term $S_c$:
\begin{equation}
  S_c = 1 - \frac{n_{err}}{n_{valid}}
\end{equation}
measures avoidance of structurally unjustified moves. Note that we omit \textit{invalid} guesses (no parseable answer) to ensure fair evaluation of real guessing attempts. Let $T_f$ be the number of tokens found out of $T$ required. The final score $S_{swm}$ is
\begin{equation}
  S_{swm} = \frac{T_f}{T} \times S_c
\end{equation}

\textbf{Human baseline.} Without history, $N{=}11$ participants achieve $\bar{S}_{swm}{=}0.900$ overall (ranging from $0.838$ on Text Hard to $0.957$ on Image-only Easy); with full history provided, scores rise to $\bar{S}_{swm}{=}0.981$. Full per-condition results are in Appendix~\ref{appendix:human-results}.

%% file: sections/3_3_wcst.tex
\textbf{Task.} In WCST, four option cards containing shapes are shown to participants. They are then asked to match the given card to one of the four option cards using one of three rules: shape, color, or number. Importantly, participants are not informed of the correct rule for matching the cards and must eliminate potential rules to find the correct one. After choosing the correct answer, a different card is shown while maintaining the same rule. The matching rule is considered complete after a fixed number of consecutive correct answers and is changed upon completion without expressly informing the participant. The test finishes after a fixed number of rules are completed or a maximum number of guesses is reached.

\textbf{Setup.} We require 5 consecutive correct matches ($c=5$) to complete a rule instance. We define two difficulty levels. \textbf{Easy}: 3 candidate rules (color, shape, number), each repeated twice (6 total rule instances) or until the 64-guess cap. \textbf{Hard}: adds background color (4 candidate rules, 8 total rule instances) with a maximum of 96 guesses. We also construct controlled ambiguity variants for the hard setting: (i) \emph{off} (no ambiguous cards), (ii) \emph{first} (only the first card under a new rule is ambiguous), and (iii) \emph{rest} (all subsequent cards after the first are ambiguous). Ambiguity occurs when multiple attributes match simultaneously, forcing the model to track disambiguating feedback over turns. The image-input setting provides images of the cards (Figure \ref{fig:neuro-cognition-task}), while the text-only setting lists descriptions of the option cards.

\textbf{Construct adaptation.} WCST primarily probes \textit{cognitive flexibility} \citep{Diamond2013}, the ability to detect rule changes from feedback and reallocate to a new active hypothesis \citep{monsell2003task}. Providing conversation history does not weaken this construct because the rule itself is a latent state that is never present in the history, so the model must still infer it from feedback patterns and re-infer it whenever feedback indicates a change.

\textbf{Metrics.} Each completed rule $i$ uses $g_i$ guesses from the start of that rule until $c$ consecutive correct response. Therefore, we can define a score $S_{r_i}$ for each completed rule $i$:
\begin{equation}
  S_{r_i} = \frac{c}{g_i}.
\end{equation}
where unfinished rules contribute zero. Lower $g_i$ means faster hypothesis convergence, which repeated reliance on an incorrect hypothesis inflates $g_i$ and lowers $S_{r_i}$. With $n_c$ completed out of $N$ target rule instances, we can calculate the final score $S_{wcst}$ as:
\begin{equation}
  S_{wcst} = \frac{1}{N}\sum_{i=1}^{n_c} S_{r_i}.
\end{equation}
We additionally report two error rates over a trial with total turns $L$, where $y_t\in\{0,1\}$ is correctness at turn $t$ and $r_t\in\mathcal{R}$ is the rule the model is following. \textbf{Perseverative Response (PR)} is the rate of re-selecting a rule already ruled out by feedback in the current block (a broader form of the standard perseverative error, PE). With $\mathcal{R}_{\neg}^t$ the set of eliminated rules at turn $t$:
\begin{equation}
  \mathrm{PR}
  =
  \frac{\sum_{t=1}^{L} \mathbf{1}\{\mathcal{R}_{\neg}^t \neq \emptyset\}\,\mathbf{1}\{r_t \in \mathcal{R}_{\neg}^{t}\}}{\sum_{t=1}^{L} \mathbf{1}\{\mathcal{R}_{\neg}^t \neq \emptyset\}}.
\end{equation}
\textbf{Failure-to-maintain-set (FMS)} marks a rule as \emph{acquired} after $m{=}3$ consecutive correct responses within a block, and counts errors on post-acquisition turns $\mathcal{T}_{\mathrm{post}} = \bigcup_{k} \{t : k(t)=k,\ t>\tau_k\}$ where $\tau_k$ is the first acquisition turn in block $k$:
\begin{equation}
  \mathrm{FMS} = \frac{\sum_{t \in \mathcal{T}_{\mathrm{post}}} \mathbf{1}\{y_t=0\}}{|\mathcal{T}_{\mathrm{post}}|}.
\end{equation}

\textbf{Human baseline.} $N{=}11$ participants achieve $\bar{S}_{wcst}{=}0.782$ without history (ranging from $0.737$ on Image Hard to $0.815$ on Text Easy) and $\bar{S}_{wcst}{=}0.838$ with history. Full per-condition results are in Appendix~\ref{appendix:human-results}.

%% file: sections/3_4_experiment_setup.tex
In our experiments, we benchmark several recent multimodal reasoning models: proprietary models (Gemini 3 Pro \citep{google2025gemini3pro}, Gemini 2.5 Pro \citep{comanici2025gemini}, GPT-5 \citep{openai2025gpt5systemcard}, o4-Mini \citep{openai2025o4minisystemcard}, Claude Sonnet 4 \citep{anthropic2024claudesonnet4systemcard}, Grok 4.1 Fast \citep{xai2025grok41fast}, Grok 4 Fast \citep{xai2025grok4fast}) and open-weight models (GLM 4.6V \citep{zai2025glm46v}, GLM 4.5V \citep{vteam2025glm45vglm41vthinkingversatilemultimodal}, Qwen3-VL-235B \cite{Qwen3-VL, Qwen2.5-VL, Qwen2VL, Qwen-VL}).

We use models' default reasoning capabilities via API parameters and chain-of-thought prompts. However, we disable reasoning for Claude Sonnet 4 and Grok 4 Fast on the RAPM test, due to overthinking issues that exhausted output limits and led to worse performance. For RAPM, we set thinking budgets close to the model's maximum output tokens. For SWM and WCST, we limit the output tokens to 8,192 tokens and the thinking budget to 4,000 tokens per turn. We repeat all SWM and WCST experiments three times, while RAPM is evaluated once. All evaluations are zero-shot with further details are in Appendix~\ref{appendix:models-experimental-setup}.
% except for the SWM image and image+text hard settings, which we run only once due to the large number of image inputs.

%% file: sections/4_1_key_findings.tex
\begin{figure}[t]
  \centering
  \includegraphics[width=\linewidth]{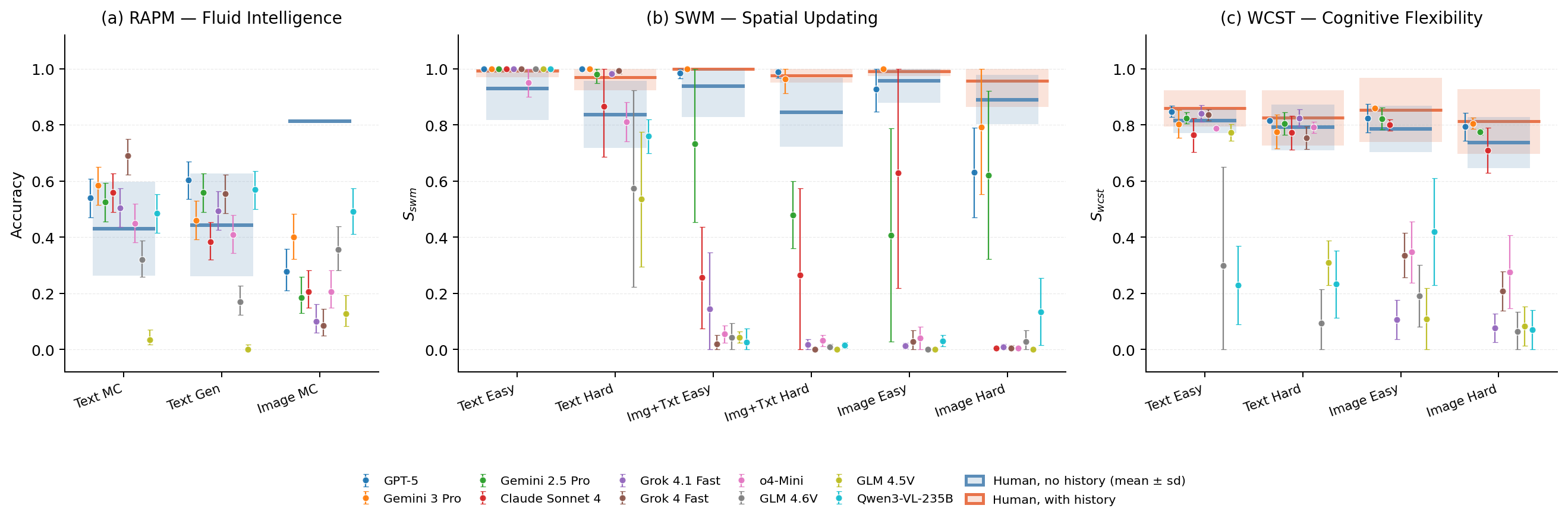}
  \caption{Human and LLM performance across the three constructs. Dots are individual models with vertical error bars showing $\pm$std over 3 runs (SWM, WCST) or 95\% CI (RAPM, single-pass with $n{=}200$ text and $n{=}140$ image). Strips show the human band (mean$\,\pm\,$sd, $N{=}11$ for SWM/WCST, $N{=}9$ for RAPM text).}
  % Full numerical results are in Table~\ref{tab:main-results}.
  \label{fig:human-vs-llm}
\end{figure}

\textbf{Models lean on text and struggle on image, while humans do not.} Across all three tests, models do better on text than image and image+text rarely closes the gap (Figure~\ref{fig:human-vs-llm}; RAPM $p{=}.010$, SWM $p{=}.001$, WCST $p{=}.037$), and difficulty also degrades scores ($p{<}.001$). The human pilot shows the opposite pattern on RAPM, where participants found the symbolic text format harder than the figural image format ($0.430/0.444$ for text MC/Gen vs.\ $0.814$ for image \citep{zhang2019raven}), and modality has no significant effect on human SWM ($p{=}.119$) while difficulty does ($p{<}.001$, $r{=}.48$). The modality penalty is therefore a property of the model rather than the task itself \citep{carroll1993human}. Consequently, on text tasks top LLMs match or exceed the human band across all three constructs, showing that the human-LLM gap is primarily modality-driven. Among all constructs, LLMs show to be strongest on WCST, where more models reach the human range across difficulties and modality, suggesting that rule switching is easier for current LLMs.

\textbf{Humans and LLMs fail at different parts of the same task.} On SWM without history, humans score $0.963$ on Image Easy and $0.878$ on Image Hard ($N{=}11$), and with history they reach near-ceiling ($\bar{S}_{swm}{=}0.982$). For humans, maintenance is the hard part. LLMs are the opposite. Even with full history in context several models drop near zero on Image Hard, and the largest jumps come from adding text on top of image (e.g., GPT-5 $0.631\to0.989$). The text adds no new state, only makes the existing state easier to reason over, so humans struggle to remember while LLMs struggle to manipulate information they already have.

\textbf{Top models are capable on image but inconsistent across runs.} Even the strongest image models are unreliable. Gemini 3 Pro scores $1.00_{\pm.00}$ on Image Easy but $0.79_{\pm.24}$ on Image Hard, and Gemini 2.5 Pro scores $0.41_{\pm.38}$ on Image Easy across three identical runs, pointing to a consistency limit rather than a capability gap. Across all ten models, SWM image scores span $0.000$ to $1.000$ while humans cluster near the top, and the model effect ($\eta^{2}{=}.88$) is far larger than the human participant effect ($\varepsilon^{2}{=}.47$).

% \textbf{Cognitive flexibility is where models do best.} WCST is the test where the most models reach the human range, on both Easy and Hard. Switching rules from feedback seems easier for current LLMs than abstract reasoning (RAPM) or planning a search (SWM). Image still pulls weaker models down sharply, which points to grounding rather than the rule-switching itself.

\textbf{Answer format matters.} In RAPM Text, the same items get different accuracies under MC and Gen, and some models swing in opposite directions (GPT-5 $0.540\to0.605$, Claude $0.560\to0.385$, Grok 4 Fast $0.690\to0.555$). MC gives a small set of choices and rewards quick elimination in trade of distraction, while Gen gives the model freedom to build an answer that meets their internal representation of the rules.
% The format itself changes how the model solves the problem, so reporting both is informative.

%% file: sections/4_2_failure_cases_patterns.tex
We split failures into \textbf{grounding failures} (basic-skill failures) and \textbf{cognitive failures} (failures of the targeted ability).
This is because simple cognitive tests still need basic skills like reading, parsing, and seeing patterns \citep{carroll1993human}, and separating those is important to understand where the models are struggling.

\label{sec:failure-rapm}
\textbf{RAPM.} From RAPM reasoning traces (Appendix \ref{sec:full-results} Table \ref{tab:rapm-full-results}, \ref{tab:rapm-failure-examples}), we identify three failure patterns. First, \textbf{hallucination and mis-grounding}, where models invent visual features or read patterns from text that are not exist. Second, \textbf{semantic over-interpretation}, where models skip simple local rules (XOR, concatenation) in favor of complex ideas like ``symmetry grids.'' Third, in Text Gen, a \textbf{content-logic disconnect}, where models find the right attribute rules but produce strings that satisfy the rules on paper while breaking the actual logic. The first is grounding, the rest are cognitive.

\label{sec:failure-swm}
\begin{figure}[t]
  \centering
  \includegraphics[width=\linewidth]{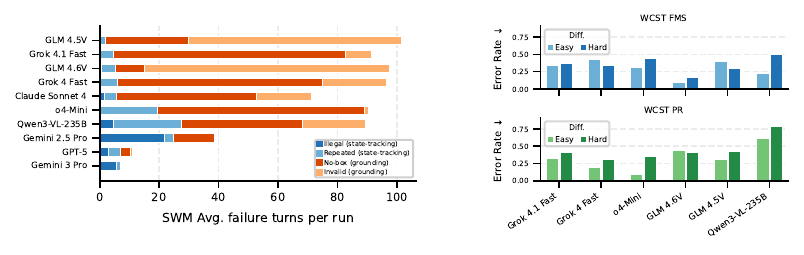}
  \caption{\textbf{Left}, SWM error mix on image and image+text averaged over difficulties, sorted by grounding failure share. \textbf{Right}, WCST FMS and PR for weaker models on image. $\downarrow$ lower is better.}
  \label{fig:swm-wcst-failures}
\end{figure}

\textbf{SWM.} On image and image+text where models struggle most (Figure~\ref{fig:swm-wcst-failures}, left), the two failure types split cleanly. \textbf{State-tracking errors} (illegal, repeated) reflect failures to manipulate state information toward the goal \citep{baddeley1974working,Diamond2013}, while \textbf{grounding errors} (no-box, invalid) reflect basic stimulus reading and instruction following. Different models fail in different ways, where top models more on state-tracking/cognitive errors while weaker models exhibit more on grounding errors.
% The two error types do not move together across models \citep{cronbach1955construct,borsboom2004concept}, which means failures can be attributed specifically to grounding or to the targeted ability rather than lumped into a single score.

\label{sec:failure-wcst}
\textbf{WCST.} Figure~\ref{fig:swm-wcst-failures} (right) shows FMS and PR for weaker models on WCST image. Most errors happen after the model has already found the right rule. The model either drifts away from the acquired rule (FMS) or goes back to a rule it already saw was wrong (PR). Adding the background-color attribute makes both worse, as expected.
% Together with the SWM split, this suggests the limit for current LLMs is not noticing that something changed, but acting on that signal reliably.

% \label{sec:failure-dissociation}
% \textbf{Why the split matters.} Read together, the benchmark behaves the way a multi-component test should. Errors split into a basic-skills family and a cognitive family, and the cognitive family further splits along the targeted component (updating in SWM, shifting in WCST). A second signal appears in Section~\ref{sec:correlation-other-benchmarks}, where RAPM Image MC correlates strongly with ARC-AGI~2 ($r{=}0.87$) but only weakly with the average of standard benchmarks ($r{=}0.09$), the pattern we expect when a test really targets a specific ability rather than general capability \citep{cronbach1955construct}.

%% file: sections/5_1_effect_reasoning_rapm.tex
\begin{figure}[t]
    \centering
    \includegraphics[width=\linewidth]{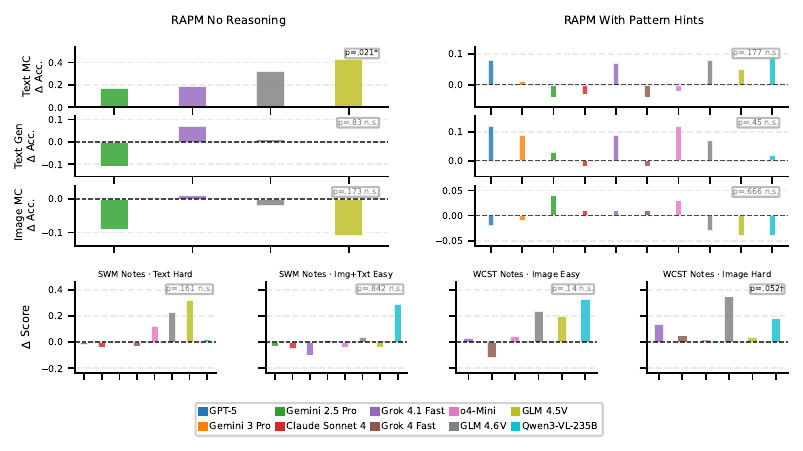}
    \caption{Ablation deltas (condition $-$ baseline). \textbf{Top}: RAPM with reasoning disabled (4 models, Text MC $p{=}.021^*$) and with pattern hints (10 models, all n.s.). \textbf{Bottom}: note-taking on SWM (left) and WCST (right); WCST Image Hard reaches $p{=}.052^{\dagger}$. Paired $t$-tests.}
    \label{fig:rapm-notes-combined}
\end{figure}

We compare RAPM with default reasoning versus a no-reasoning constraint (Figure~\ref{fig:rapm-notes-combined}, top-left) on models where reasoning can be turned off. Disabling reasoning gives a large and consistent gain in Text MC ($p=.021$), while Text Gen and Image MC are mixed ($p=.83$ and $p=.173$, n.s.). In Text MC with 8 choices, long chains of thought tend to confuse the model because of distractors. In Text Gen, the open answer space heavily benefits from the ability to build a coherent answer. In Image MC, results suggest that visual stimuli still benefit from explicit step-by-step parsing.

% In short, reasoning helps when reading the stimulus is the hard part and hurts when overthinking a small choice is the hard part.

% Ablation figure moved to Appendix~\ref{appendix:ablation-figures} to free space for the human--LLM dissociation figure in Section~\ref{sec:results}.

% Text MC has a small explicit answer set, and long chains of thought tend to drift toward distractors. In Text Gen the answer space is open, so any benefit from skipping deliberation is offset by the cost of building a coherent answer. In Image MC the trend goes the other way, suggesting that visual stimuli still benefit from explicit step-by-step parsing. In short, reasoning helps when reading the stimulus is the hard part and hurts when overthinking a small choice is the hard part.

%% file: sections/5_2_improve_pattern_rapm.tex
In RAPM, we test whether providing pattern hints helps models narrow their reasoning. The results in Figure~\ref{fig:rapm-notes-combined} (top-right) vary across models and conditions, and the average effect is not significant in any setting (Text MC $p=.177$, Text Gen $p=.45$, Image MC $p=.666$). Most cells barely move and a few drop. Hints only help if the model can act on them, so the null result suggests the gap is in the abstract reasoning step itself rather than in knowing which patterns to consider.

%% file: sections/5_3_improve_notes_swm_wcst.tex
For SWM and WCST, we provide automatic notes that emulate explicit cognitive offloading \citep{risko2016cognitive,larkin1980expert}, restating the current state and prior actions every turn. From Figure~\ref{fig:rapm-notes-combined} (bottom-left), the average effect on SWM is not significant (Text Hard $p=.161$, Img+Text Easy $p=.842$). This is what we expected, since the in-context history already covers the maintenance demand, thus adding notes did not solve the manipulation issues for LLMs.
% The failures came from the manipulation part rather than the memory load.
For WCST (Figure~\ref{fig:rapm-notes-combined}, bottom-right), notes help more consistently and approach significance under the harder image condition (Image Easy $p=.14$, Image Hard $p=.052$). Notes externalize the current rule guess and feedback, which is exactly what cognitive flexibility needs \citep{Diamond2013} and the in-context history does not provide.
% The contrast with SWM is itself useful. The same intervention helps where it adds new information (rule state in WCST) and barely helps where it duplicates what is already in context (move history in SWM).

%% file: sections/6_1_factor_analysis_llm.tex
To understand the latent structure of capabilities across widely used LLM benchmarks inspired by \citet{burnell2023revealingstructurelanguagemodel}, we conducted an exploratory factor analysis (EFA) on performance data obtained from Artificial Analysis\footnote{https://artificialanalysis.ai/leaderboards/models} for 156 distinct LLMs. We selected 10 benchmarks widely recognized as standard measures of LLM ``intelligence'' and real-world capabilities, which satisfy the large sample size ($N>100$) requirement for valid factor analysis. We used Principal Axis Factoring for the factoring method.
% ranging from coding and math to agentic tool use.
% Importantly, this selection was driven by the statistical requirement for a large sample size ($N>100$) to ensure valid factor analysis, which very few other benchmarks satisfy. We used Principal Axis Factoring for the factoring method.

Parallel analysis indicated a single-factor solution: a unidimensional latent variable explains most performance variation across these tasks. This factor (PA1) accounted for 75\% of the total variance, providing strong evidence for a "general capability" factor. As shown in Table~\ref{tab:factor_loadings} (Appendix~\ref{appendix:factor-analysis}), all benchmarks had high loadings ($\lambda > 0.76$).
% with \textit{GPQA Diamond} (0.943) and \textit{LiveCodeBench} (0.926) as the strongest proxies.
This coherence also exposes a limitation of current evaluation, where many benchmarks measure the same underlying construct. This makes NeuroCognition even more important, as it targets distinct, basic cognitive tasks.

%% file: sections/6_2_correlation_other_benchmarks.tex
To assess how NeuroCognition relates to real-world tasks, we correlate the aggregate NeuroCognition score with the average of 11 state-of-the-art benchmarks (the 10 in Section~\ref{sec:factor-analysis-llm} plus MMMU Pro \cite{yue-etal-2025-mmmu}) as a proxy for \textit{general capability}. Pearson correlations are shown in Figure~\ref{fig:corr-all} (detailed setups in Appendix~\ref{sec:external-benchmark}). We observe a high correlation between the average NeuroCognition score and the 11-benchmark average ($r{=}.86, p{=}.001, N{=}10$), validating our framework as a robust predictor of general model capability ($g$). We also correlate each test with an external benchmark that targets the same cognitive demand:

(1) \textbf{Spatial Updating over Long Context}: \textit{SWM Text Hard} correlates moderately with \textit{Long HotPotQA} \citep{yang2018hotpotqadatasetdiverseexplainable}, a modified version that splits questions and passages at the word level and treats each word as a turn ($r{=}.43, p{=}.21, N{=}10$). The moderate (rather than strong) correlation reflects different demand intensities. SWM requires long-context search, reasoning, and acting on state \textit{every turn}, whereas Long HotPotQA only requires answering a single question after reading a long passage.

(2) \textbf{Fluid Intelligence}: \textit{RAPM Image MC} is the strongest predictor of \textit{ARC-AGI 2} \citep{chollet2019measureintelligence} ($r{=}.87, p{=}.024, N{=}6$), consistent with isolating an ``abstract visual reasoning'' factor that standard benchmarks often miss, and uncorrelated with the general benchmark average ($r{=}.09$).

(3) \textbf{Cognitive Flexibility}: \textit{WCST Text Hard} shows a weak, non-significant correlation with \textit{LLM Task Interference} \citep{gupta2024llmtaskinterferenceinitial} ($r{=}.24, p{=}.50, N{=}10$). Again the gap reflects demand intensity: LLM Task Interference tests task-switching once per item, while WCST requires repeated rule re-inference whenever the rule changes within a session.

%% file: sections/6_3_relation_with_human.tex
Among the 10 models, only GPT-5 ($\rho{=}{+}0.72$, $p{=}.020$) and Gemini~3~Pro ($\rho{=}{+}0.79$, $p{=}.007$) match the human agreement range (human-human median $\rho{=}0.787$; Appendix~\ref{appendix:human-results-insights}). This shows that these two models possess human-like cognitive sensitivity, suggesting they share structure with human cognition in terms of what drives errors on these constructs. The remaining eight show near-zero or negative $\rho$, meaning they are relatively better where humans struggle and worse where humans excel. This \textit{qualitative} misalignment, rather than a uniform performance gap, is what distinguishes the majority of current models from human cognition on these tasks. 

% The two aligned models show human-like cognitive sensitivity: they respond to task difficulty and modality in the same direction and magnitude as humans, suggesting they engage with the same features humans find demanding. The eight misaligned models, by contrast, appear to solve these tasks through a qualitatively different process --- one that does not share the same cognitive bottlenecks.

%% file: sections/7_conclusion.tex
In this work, we introduced the NeuroCognition benchmark, a suite of three neuropsychological tests (RAPM, SWM, WCST) adapted for LLMs to probe specific cognitive processes using multimodal, process-aware metrics. Across tasks, state-of-the-art models show meaningful but incomplete alignment with human cognition. They perform better in text, but performance and consistency drops sharply for images and harder settings that demand intensive grounding and goal-directed spatial updating of task state. Comparison with human baseline shows a construct-level dissociation. Within tasks, humans struggle to maintain task state while LLMs struggle to manipulate information they already have in context.  Ablation study showed that explicit reasoning is not uniformly beneficial, and human-like assistance shows mixed effects. Complementing these results, our factor analysis suggests a dominant general-capability factor across standard benchmarks, highlighting redundancy in current evaluation. Correlations with external benchmarks further suggest that gains in NeuroCognition process scores translate into improvements in semantically related real-world tasks, while individual tests still measure distinct cognitive abilities beyond the shared general factor. Overall, our results suggest that current LLMs do not possess the same kind of intelligence as humans, but align with humans in specific aspects, and the patterns of misalignment are themselves diagnostic, indicating which cognitive components current models still need to develop.

% The human baseline reveals a striking construct-level dissociation. Within each test, humans approach the practical ceiling, in the order neuropsychology predicts (fluid intelligence and goal-directed manipulation of working memory before sustained cognitive flexibility under uncertain feedback), while LLMs invert this ordering. Strong models match the human band on cognitive flexibility (WCST) yet collapse on SWM-image, the construct humans find easiest. Within tasks the failure modes also differ: humans struggle to maintain task state, whereas LLMs struggle to manipulate information they already have in context, and the post-acquisition WCST error pattern (FMS, PR) seen in the model traces is rarely observed in human protocols. Together with the modality penalty (model effect $\eta^{2}{=}.88$ vs.\ human participant effect $\varepsilon^{2}{=}.47$), these patterns argue that scalar parity on a benchmark cell is a weak form of similarity and does not imply a shared cognitive substrate.

%% file: appendix/human_protocol.tex
\section{Human Experiment Protocol}
\label{appendix:human-protocol}

\textbf{Participants and recruitment.} $N{=}11$ participants completed the SWM and WCST tasks under both history conditions, and $N{=}9$ participants completed the RAPM text pilot ($N{=}9$ MC, $N{=}9$ Gen). Participants were recruited via an open broadcast targeting adults aged 17 or older who are physically, mentally, and psychologically healthy. No prior experience with cognitive testing was required. Demographics are reported in Appendix~\ref{appendix:human-results}.

\textbf{Procedure.} Each session followed a fixed sequence: (1) the researcher provided an overview of the task using the participant guidelines (Section~\ref{appendix:human-guidelines} below); (2) participants provided written informed consent before any testing began; (3) participants were given the opportunity to ask clarifying questions; (4) the task was conducted; (5) participants completed a post-test questionnaire. Conditions were completed in separate sessions to avoid carry-over.

\textbf{Conditions.} Each participant completed SWM and WCST in two conditions:
\begin{itemize}[itemsep=0.15em, topsep=0.15em]
  \item \textbf{Without history.} Original CANTAB-style protocol. No prior moves or feedback are visible, and the participant must maintain the task state (which boxes have been opened, which rule the last feedback eliminated) entirely in working memory. This is the standard reference condition for the human baseline.
  \item \textbf{With history.} Full move and feedback history is shown to the participant on every turn, mirroring the in-context history that LLMs always receive in our evaluation. This isolates the manipulation demand by externalising the maintenance demand.
\end{itemize}
% LLMs are stateless and require state to be supplied in context, so the with-history setting is the only LLM evaluation setup. The without-history human condition is reported as a calibration reference rather than a matched LLM comparison, and the with-history human strip in Figure~\ref{fig:human-vs-llm} provides the matched comparison.

\textbf{Sessions.} SWM conditions: Text Easy/Hard, Image+Text Easy/Hard, Image-only Easy/Hard (3 runs each for the no-history condition, 1 run for history). WCST conditions: Text Easy/Hard, Image Easy/Hard (1 run per condition per history setting). RAPM text was conducted as two separate sessions per participant: Multiple-Choice (MC, 15 items) and Generative (Gen, 15 items). RAPM has no with-history variant since each item is a single-turn completion.

\textbf{Scoring.} Identical metrics ($S_{swm}$, $S_{wcst}$, error rates) were computed for human responses. Since human participants only provide invalid outputs unintentionally (mistype, etc.), the invalid error type does not apply. We report the human-adjusted $S_{swm}$ (which excludes no-box events from the denominator) as the primary metric to maintain consistency with the model evaluation. Run-to-run reliability for the no-history SWM condition (3 runs per participant$\times$condition, $N{=}66$ units) was assessed via ICC(3,1): $\mathrm{ICC}(3,1){=}0.738$, indicating good reliability (standard threshold $>{0.75}$) and confirming that participant and condition effects dominate over run-to-run noise.

\textbf{Ethics.} The study involves no deception, no sensitive data collection, and no physical or psychological risk beyond normal cognitive task engagement. Participants provided informed consent. The study was conducted in accordance with the ethical guidelines of the authors' institution. Participants received financial compensation according to the local minimum hourly wage.

\subsection{Participant Guidelines}
\label{appendix:human-guidelines}

The following guidelines were provided to all participants before the study. Task instruction images shown during the session are omitted here for brevity; the full guidelines with screenshots will be released alongside the code and data.

\subsubsection*{Overview}
Participants completed one or more of the following tasks: Spatial Working Memory (SWM), Wisconsin Card Sorting Test (WCST), and Raven's Advanced Progressive Matrices (RAPM). Participants were asked to focus on solving the task based on what they saw and the feedback they received.

\subsubsection*{SWM Task Instructions}
\textbf{Objective.} Find token(s) hidden in boxes by opening them one at a time. When a token is found, it regenerates in a box that has not yet contained that token. The test continues until all tokens have been found in every box.

\textbf{Scoring.} The score starts at 1.0 and is reduced for each illegal move (opening a box that cannot contain a token) or repeated move (opening a known-empty box). Speed is not a factor.

\textbf{How to answer.}
\begin{itemize}[itemsep=0.1em, topsep=0.1em]
  \item \textit{Text setup}: enter a single box number (boxes numbered 1--8 or 1--12).
  \item \textit{Image/image+text setup}: enter coordinates as \texttt{col,row} (0-indexed).
\end{itemize}

\textbf{Key rules.}
\begin{itemize}[itemsep=0.1em, topsep=0.1em]
  \item A box may contain multiple token types simultaneously.
  \item Once a token type is found in a box, it will not reappear there.
  \item Taking notes is prohibited. Use the provided history (when enabled).
\end{itemize}

\textbf{Difficulty levels.} \textit{Easy}: 8 boxes, 1 token type. \textit{Hard}: 12 boxes, 2 token types. Setups: text, image+text, and image-only.

\subsubsection*{WCST Task Instructions}
\textbf{Objective.} Match a given card to one of four option cards using a hidden sorting rule (color, shape, or number of shapes). After each response you receive correct/incorrect feedback. The rule changes periodically, so you need to adapt accordingly.

\textbf{How to answer.} Enter a single number: 1, 2, 3, or 4.

\textbf{Key rules.}
\begin{itemize}[itemsep=0.1em, topsep=0.1em]
  \item Identify the current matching rule, then maintain and adapt based on feedback.
  \item Taking notes is prohibited.
\end{itemize}

\textbf{Difficulty levels.} \textit{Easy}: 3 candidate rules, no background-color attribute, 64-trial limit. \textit{Hard}: 4 candidate rules including background color, 100-trial limit. Setups: text and image+text.

\subsubsection*{RAPM Task Instructions}
\textbf{Objective.} Analyze a $3{\times}3$ matrix of symbolic and string patterns and identify the rule governing each row and column. Select or generate the entry that completes the missing bottom-right cell.

\textbf{Scoring.} One point per correct answer out of 15 items.

\textbf{How to answer.}
\begin{itemize}[itemsep=0.1em, topsep=0.1em]
  \item \textit{Multiple-choice (MC)}: enter the number of the correct option (1--8).
  \item \textit{Generative (Gen)}: enter the string that represents the correct pattern.
\end{itemize}

\textbf{Key rules.} Taking notes is allowed. AI assistance and internet search are prohibited. Participants completed 15 MC items and 15 Gen items in separate sessions.

%% file: appendix/human_results.tex
\section{Human Experiment Results}
\label{appendix:human-results}

% We report human SWM and WCST performance under both conditions described in Appendix~\ref{appendix:human-protocol}, the RAPM text pilot, cross-cutting insights, and post-test questionnaire highlights. The human-LLM dissociation is summarised in Figure~\ref{fig:human-vs-llm}, with full numerical results in Table~\ref{tab:main-results}.

\subsection{SWM and WCST}
\label{appendix:human-results-swm-wcst}

% Figure~\ref{fig:human-consolidated} reports paired boxplots for every modality $\times$ difficulty condition under both history settings, with run counts and per-condition Mann--Whitney significance markers. Table~\ref{tab:human-consolidated} reports the corresponding numerical means and per-condition $\Delta$.

\begin{table}[H]
  \centering
  \small
  \caption{Human SWM (human-adjusted) and WCST per-condition means$_{\pm\text{std}}$, no-hist run count, and with-hist run count. $\Delta$ is the with-hist mean minus the no-hist mean.}
  \label{tab:human-consolidated}
  \begin{tabular}{llrrr}
    \toprule
    \textbf{Task} & \textbf{Condition} & \textbf{No-hist mean$_{\pm\text{sd}}$ ($n$)} & \textbf{With-hist mean$_{\pm\text{sd}}$ ($n$)} & \textbf{$\Delta$} \\
    \midrule
    \multirow{6}{*}{SWM}
    & Text Easy        & $0.929_{\pm.11}$ (33) & $0.993_{\pm.02}$ (11) & $+0.064$ \\
    & Text Hard        & $0.838_{\pm.12}$ (33) & $0.970_{\pm.05}$ (11) & $+0.132$ \\
    & Image+Text Easy  & $0.938_{\pm.11}$ (33) & $1.000_{\pm.00}$ (11) & $+0.062$ \\
    & Image+Text Hard  & $0.845_{\pm.13}$ (33) & $0.976_{\pm.03}$ (11) & $+0.131$ \\
    & Image-only Easy  & $0.957_{\pm.08}$ (33) & $0.992_{\pm.02}$ (11) & $+0.035$ \\
    & Image-only Hard  & $0.891_{\pm.09}$ (33) & $0.957_{\pm.10}$ (11) & $+0.066$ \\
    \midrule
    \multirow{4}{*}{WCST}
    & Text Easy        & $0.815_{\pm.05}$ (11) & $0.859_{\pm.07}$ (11) & $+0.044$ \\
    & Text Hard        & $0.792_{\pm.09}$ (11) & $0.825_{\pm.10}$ (11) & $+0.033$ \\
    & Image Easy       & $0.786_{\pm.09}$ (11) & $0.855_{\pm.12}$ (11) & $+0.069$ \\
    & Image Hard       & $0.737_{\pm.10}$ (11) & $0.813_{\pm.12}$ (11) & $+0.076$ \\
    \bottomrule
  \end{tabular}
\end{table}

\begin{figure}[H]
  \centering
  \includegraphics[width=\linewidth]{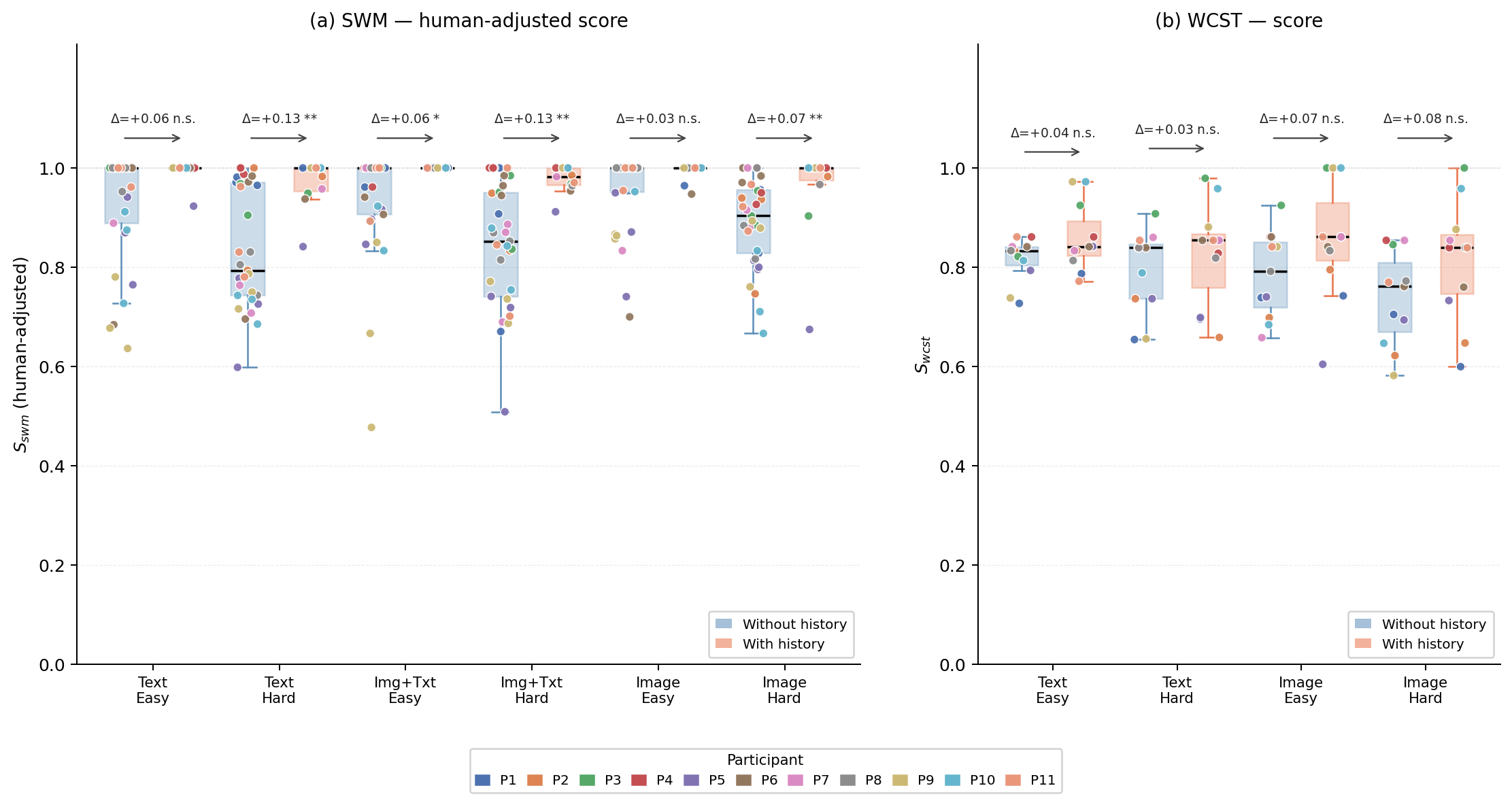}
  \caption{Human SWM (a) and WCST (b) per-condition performance under without-history (blue) and with-history (orange). $\Delta$ = mean difference, stars = Mann--Whitney $p$ ($^{*}<.05$, $^{**}<.01$, $^{***}<.001$).}
  % Aggregate effects: SWM $0.900 \to 0.981$, $\Delta{=}{+}0.082$, $p{<}.001$, $r{=}0.43$ ($N{=}11$ each). WCST $0.782 \to 0.838$, $\Delta{=}{+}0.056$, $p{=}.011$, $r{=}0.32$ ($N{=}11$ each).}
  \label{fig:human-consolidated}
\end{figure}

\textbf{SWM history effect.} Providing full move history drives human SWM scores nearly to ceiling ($0.900 \to 0.981$, $p{<}.001$, $r{=}0.43$). The largest gains are on Hard conditions (up to $+0.132$ on Text Hard and $+0.131$ on Image+Text Hard), where maintaining state across many tokens is most demanding. Without history the bottleneck for humans is memory \textit{maintenance}, not spatial \textit{manipulation} itself. With history, the maintenance demand is externalised and near-perfect performance becomes achievable. This is precisely the construct decomposition our SWM adaptation (Section~\ref{sec:swm}) is designed to enable, by isolating manipulation demand from maintenance demand.

\textbf{WCST history effect.} WCST also shows a small but significant history benefit ($+0.056$, $p{=}.011$, $r{=}0.32$). The matching rule is a latent variable that is never present in the history, so participants must still infer it from feedback regardless of whether prior turns are visible, and the benefit is correspondingly modest. The size of the SWM gain ($+0.082$) is roughly $1.5{\times}$ the WCST gain, and the effect-size $r$ is also larger ($0.43$ vs.\ $0.32$), so history aids WCST less than it aids SWM. This is consistent with the construct separation between the two tasks. SWM is largely maintenance-bound for humans whereas WCST is largely rule-inference-bound, with history helping the latter only insofar as it reduces lookup cost on prior feedback.

\label{appendix:human-results-insights}
\textbf{Modality, difficulty, and participant effect.} Across SWM and WCST, human performance is shaped mainly by difficulty rather than modality. In SWM, modality has no significant effect ($p{=}.119$) but difficulty does ($p{<}.001$, $r{=}.48$), while in WCST the text-over-image advantage is small (Text Easy: $0.815$, Image Easy: $0.786$), much smaller than the image penalty seen in many LLMs. Humans also vary less than models do. The participant effect is significant ($\varepsilon^{2}{=}.47$) but smaller than the LLM model effect ($\eta^{2}{=}.88$). This variation is mostly quantitative, not qualitative, since participants largely agree on which conditions are harder (median pairwise Spearman $\rho{=}0.787$), show good within-participant run-to-run reliability ($\mathrm{ICC}(3,1){=}0.738$), and mainly attribute errors to forgetting previously opened boxes rather than misunderstanding the task.

\subsection{RAPM Text}
\label{appendix:human-results-rapm}

\begin{figure}[H]
  \centering
  \includegraphics[width=0.45\linewidth]{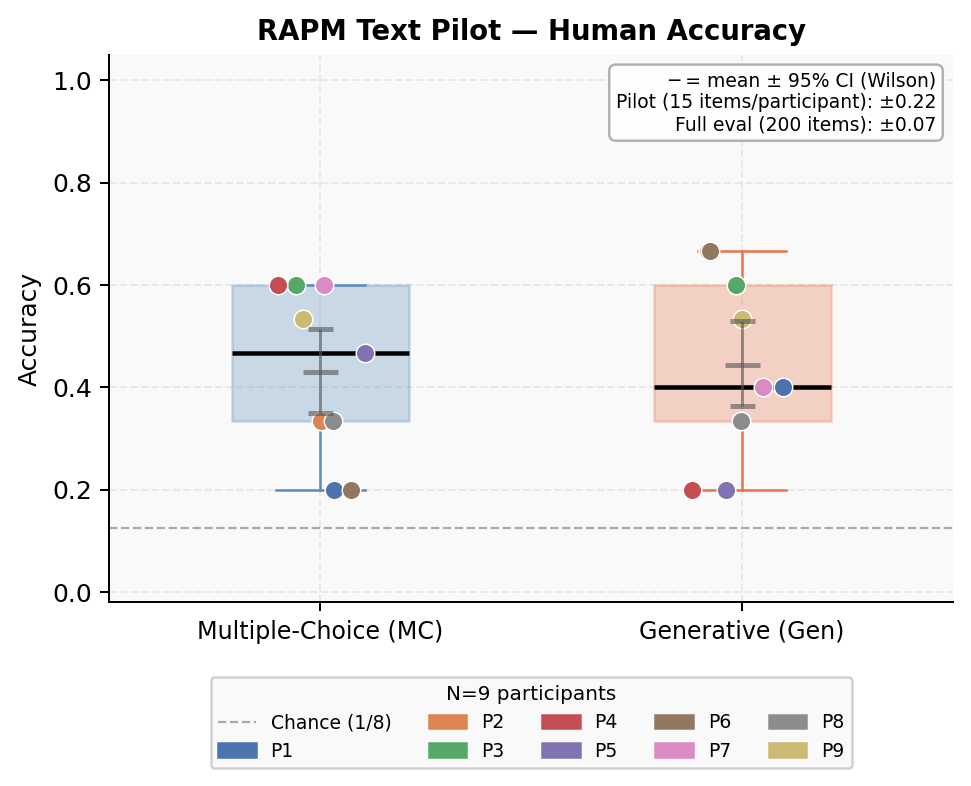}
  \caption{RAPM text pilot accuracy ($N{=}9$ participants, each completing both MC and Gen). Crossbar shows mean $\pm$ 95\% CI (Wilson). Dashed line marks random chance ($1/8$, applicable to MC).}
  % CI half-widths: pilot ($n{=}15$ items/participant) $\approx{\pm}0.25$; full LLM eval ($n{=}200$) $\approx{\pm}0.07$.}
  \label{fig:human-rapm-text}
\end{figure}

Overall, mean accuracy is similar in MC and Gen, and both exceed random chance ($1/8 = 0.125$), confirming that the items are solvable but non-trivial. Participants span a broad range of scores without approaching ceiling, which suggests that the items are genuinely discriminative. At the individual level, some participants score higher on MC and others on Gen, with no consistent direction, so answer format appears to change how humans solve the items rather than uniformly helping or hurting. This contrasts with LLMs, which show much larger format differences, indicating that format sensitivity is more a model property than a task property (Figure~\ref{fig:human-vs-llm}).

\subsection{Post-Test Questionnaire}
\label{appendix:post-test}

% Participants completed a post-test questionnaire covering demographics, difficulty, and strategies used.

\textbf{Demographics.} SWM/WCST participants ($N{=}11$) were young adults aged 18--29 (mean~25), predominantly male (7M/4F), with education ranging from high school to doctorate and moderate-to-high prior familiarity with cognitive tasks (mean~3.4/5). RAPM participants ($N{=}9$) were similarly aged 18--31 (mean~26) with comparable education and familiarity (mean~3.3/5).

\textbf{Notable subjective findings.} The clearest result from the questionnaire is that all SWM participants found the with-history condition substantially easier. This matches the near-ceiling performance in that condition. For SWM errors, all respondents identified memory failure, namely forgetting the state of the boxes, as the primary source of error rather than the spatial manipulation itself. This supports the construct decomposition, where human struggle in maintenance rather than manipulation. Strategy reports also help explain the significant participant effect in Appendix~B.1. Participants who described active planning strategies, especially for hard SWM, tended to perform better than those who relied on simple memorization or reported no clear strategy. Examples include assigning boxes numbers instead of remembering positions in the image condition, and isolating one token at a time in the two-token setup to reduce memory load. For RAPM, reported mental effort was very high (mean~6.5/7), consistent with the low accuracy. All respondents also preferred the MC format over the generative format, while mean results show similar numbers. For WCST, participants found it relatively easy to incorporate feedback (mean~2.1/5), in line with the high WCST scores, and most found the additional background-color attribute at least mildly harder.

%% file: appendix/rapm_text_details.tex
\section{RAPM Text Data Generation}
\label{appendix:text-rapm-generation}

% Tighter list spacing for this appendix (requires enumitem; already used below).
\setlist[itemize]{itemsep=0.15em, topsep=0.15em, parsep=0pt, partopsep=0pt}
\setlist[enumerate]{itemsep=0.15em, topsep=0.15em, parsep=0pt, partopsep=0pt}

% The text-based RAPM dataset is programmatically generated by constructing 3x3 matrices where each cell contains a string sequence governed by latent rules along the rows and columns.
% to evaluate abstract reasoning over symbolic string patterns. Unlike the image-based version which uses fixed visual patterns, the text version

\subsection{Rule Space and Attributes}
The generation engine employs a compositional rule space defined by attributes that constrain string properties. For each matrix, two primary attributes are sampled—one for rows and one for columns. Additional consistency rules (leaks) are inferred if they naturally emerge from the intersection of constraints.

The implemented attributes and their parameters are:

\begin{enumerate}[leftmargin=*]
    \item \textbf{Character Set Type}: Restricts the alphabet of the string.
    \begin{itemize}
        \item \textit{Values}: \texttt{letters}, \texttt{digits}, \texttt{symbols}.
        \item \textit{Logic}: All characters in the cell must belong to the specified set.
    \end{itemize}

    \item \textbf{Type Count}: Constraints on the count of specific character types.
    \begin{itemize}
        \item \textit{Types}: vowels, consonants, uppercase/lowercase variants, etc.
        \item \textit{Rules}: \texttt{even}, \texttt{odd}, \texttt{multiple\_of\_N} (for $N \in \{2, 3, 4\}$).
        \item \textit{Logic}: The count of characters matching the type must satisfy the parity or divisibility rule. Zero counts are explicitly disallowed for these rules to ensure pattern salience.
    \end{itemize}

    \item \textbf{Quantitative Constant}: Fixes a metric to a specific integer value across the sequence.
    \begin{itemize}
        \item \textit{Metrics}: String length, unique character count, or count of specific types.
        \item \textit{Range}: Integers in [2, 5].
    \end{itemize}

    \item \textbf{Quantitative Progression}: Enforces an arithmetic progression on a metric.
    \begin{itemize}
        \item \textit{Metrics}: Same as Quantitative Constant.
        \item \textit{Parameters}: Start value $\in [1, 3]$, Step size $\in [1, 3]$.
        \item \textit{Logic}: The metric value increases by the step size across the row or column (e.g., lengths 3, 5, 7).
    \end{itemize}

    \item \textbf{Sorted Order}: Enforces lexicographical ordering.
    \begin{itemize}
        \item \textit{Values}: \texttt{ascending}, \texttt{descending}, \texttt{mixed}.
        \item \textit{Logic}: Characters in the string must follow the specified order. \texttt{mixed} explicitly requires the string to be unsorted.
    \end{itemize}

    \item \textbf{Positional Constraint}: Fixes the character type at specific indices.
    \begin{itemize}
        \item \textit{Indices}: \texttt{first}, \texttt{last}, \texttt{even}, \texttt{odd}.
        \item \textit{Logic}: Characters at the specified positions must belong to a specific type (e.g., "first character must be a digit").
    \end{itemize}
\end{enumerate}

\subsection{Generation Algorithm}
The matrix generation process follows a constraint-satisfaction approach with backtracking, as detailed in Algorithm \ref{alg:rapm_gen}. The core logic proceeds as follows:

\begin{enumerate}
    \item \textbf{Attribute Sampling}: Randomly sample one attribute specification for rows and one for columns.
    \item \textbf{Compatibility Check}: Validate that the row and column attributes are compatible. For example, a "Sorted" row attribute conflicts with a "Positional" column attribute because sorting rearranges characters, potentially violating fixed-position constraints. Infeasible combinations trigger a re-sampling.
    \item \textbf{Constraint Propagation}: The row and column attributes are combined to define specific constraints for each of the $3 \times 3$ cells. "Leak" detection is performed to identify additional properties that effectively become constant across rows or columns (e.g., if a row rule restricts characters to digits and a column rule restricts them to even numbers, a universal "digits" constraint might apply).
    \item \textbf{Grid Construction}: Cells are generated sequentially. For each cell, a string is constructed to satisfy the merged constraints. Random backtracking is used if a cell cannot be generated given the current constraints and history.
    \item \textbf{Variation Enforcement}: To prevent trivial patterns, we enforce a minimum Hamming distance between strings in the same row or column. This ensures that the progression relies on abstract rules rather than simple repetition.
\end{enumerate}

Upon successful generation, the defining constraints for the missing cell (including ``leaked'' rules) are stored for automatic rule validation.

\begin{algorithm}[t]
   \caption{RAPM Text Matrix Generation}
   \label{alg:rapm_gen}
\begin{algorithmic}
   \STATE {\bfseries Input:} Random Seed $\mathcal{S}$
   \STATE {\bfseries Output:} $3 \times 3$ String Matrix $\mathcal{M}$, Constraints $\mathcal{C}$
   \STATE Initialize RNG with $\mathcal{S}$
   \REPEAT
   \STATE Sample row attribute $\mathcal{A}_{row}$ and column attribute $\mathcal{A}_{col}$
   \STATE $valid \leftarrow \text{CheckCompatibility}(\mathcal{A}_{row}, \mathcal{A}_{col})$
   \UNTIL{$valid$}
   \STATE Initialize cell constraint grid $\mathcal{C}_{3 \times 3}$ from $\mathcal{A}_{row}$ and $\mathcal{A}_{col}$
   \STATE Apply inferred consistency rules (leaks) to $\mathcal{C}$
   \FOR{$i=0$ {\bfseries to} $2$}
   \FOR{$j=0$ {\bfseries to} $2$}
   \STATE $attempts \leftarrow 0$
   \REPEAT
   \STATE Generate string $s$ satisfying $\mathcal{C}_{i,j}$
   \STATE $v_{row} \leftarrow \forall k < j: \text{Hamming}(s, \mathcal{M}_{i,k}) \ge 2$
   \STATE $v_{col} \leftarrow \forall k < i: \text{Hamming}(s, \mathcal{M}_{k,j}) \ge 2$
   \STATE $attempts \leftarrow attempts + 1$
   \UNTIL{$(v_{row} \land v_{col})$ {\bfseries or} $attempts > MAX\_RETRIES$}
   \IF{not $(v_{row} \land v_{col})$}
   \STATE \textbf{fail} and \textbf{backtrack}
   \ENDIF
   \STATE $\mathcal{M}_{i,j} \leftarrow s$
   \ENDFOR
   \ENDFOR
   \STATE \textbf{return} $\mathcal{M}, \mathcal{C}$
\end{algorithmic}
\end{algorithm}

\subsection{Multiple-Choice and Distractor Generation}
For the text MC setup, we generate 8 options: 1 correct answer (ground truth for the bottom-right cell) and 7 distractors. Each distractor is guaranteed to violate at least one latent rule to ensure a unique solution.

\begin{itemize}
    \item \textbf{Correct Answer}: The generated string for the bottom-right ($3,3$) cell.
    \item \textbf{Distractors}: Generated by applying mutation strategies to the correct answer to violate latent rules. Strategies include:
    \begin{itemize}
        \item \textit{Break Ordering}: Swapping adjacent characters to violate sort constraints.
        \item \textit{Break Positional}: Changing a character at a critical index (e.g., first/last) to a forbidden type.
        \item \textit{Adjust Count}: Adding or removing characters to violate parity or length constraints.
        \item \textit{Character Mutation}: Randomly altering characters to valid but incorrect values.
    \end{itemize}
\end{itemize}

\newpage
\section{RAPM Text Gen Automatic Rule Validation}
\label{appendix:text-rapm-validation}

For the text Gen setup, we can programmatically check if a generated string $S$ satisfies the stored underlying constraints $\mathcal{C}$ for the missing cell $(3,3)$.  Algorithm \ref{alg:cell_satisfies} details this validation procedure.

\begin{algorithm}[h!]
   \caption{RAPM Text Gen Validation (CellSatisfies)}
   \label{alg:cell_satisfies}
\begin{algorithmic}
   \STATE {\bfseries Input:} Candidate String $S$, Constraints $\mathcal{C}$
   \STATE {\bfseries Output:} Boolean (True if valid)
   \IF{$\mathcal{C}.fixed\_length \neq \text{None}$ \AND $|S| \neq \mathcal{C}.fixed\_length$}
       \STATE \textbf{return} False
   \ENDIF
   \FORALL{$(type, count) \in \mathcal{C}.target\_counts$}
       \IF{$\text{Count}(S, type) \neq count$} \STATE \textbf{return} False \ENDIF
   \ENDFOR
   \FORALL{$(type, rule) \in \mathcal{C}.parity\_rules$}
       \STATE $c \leftarrow \text{Count}(S, type)$
       \IF{$c = 0$} \STATE \textbf{return} False \COMMENT{Zero count forbidden for parity} \ENDIF
       \IF{($rule = \text{'even'} \land c \text{ is odd}$) \OR ($rule = \text{'odd'} \land c \text{ is even}$)}
           \STATE \textbf{return} False
       \ENDIF
   \ENDFOR
   \FORALL{$(type, k) \in \mathcal{C}.multiple\_rules$}
       \STATE $c \leftarrow \text{Count}(S, type)$
       \IF{$c = 0$ \OR $c \pmod k \neq 0$} \STATE \textbf{return} False \ENDIF
   \ENDFOR
   \IF{$\mathcal{C}.unique\_count \neq \text{None}$ \AND $|\text{Unique}(S)| \neq \mathcal{C}.unique\_count$}
       \STATE \textbf{return} False
   \ENDIF
   \IF{$\mathcal{C}.ordering \neq \text{None}$ \AND $\text{CheckOrder}(S, \mathcal{C}.ordering) = \text{False}$}
       \STATE \textbf{return} False
   \ENDIF
   \IF{$\mathcal{C}.positional \neq \text{None}$ \AND $\text{CheckPositional}(S, \mathcal{C}.positional) = \text{False}$}
       \STATE \textbf{return} False
   \ENDIF
   \STATE \textbf{return} True
\end{algorithmic}
\end{algorithm}

The validation function \texttt{CellSatisfies}(S, $\mathcal{C}_{3,3}$) performs the following checks:

\begin{enumerate}[itemsep=0.1em, topsep=0.1em]
    \item \textbf{Length Check}: If a fixed length is enforced (from quantitative constraints), $|S|$ must match.
    \item \textbf{Type Counts}: $Count(S, \text{type}) = k$ for all exact count constraints.
    \item \textbf{Parity/Divisibility}: $Count(S, \text{type})$ must satisfy even/odd/multiple conditions. Crucially, as enforced during generation, zero counts are invalid for these rules to prevent trivial satisfactions (e.g., "0 vowels" satisfying "even vowels").
    \item \textbf{Unique Characters}: If unique count is constrained, $|\text{Unique}(S)|$ is checked against the target.
    \item \textbf{Ordering}: If sorted order is required, $S$ is checked against its own sorted version.
    \item \textbf{Positional}: Characters at constrained indices (first, last, even, odd) are verified against the allowed character set.
\end{enumerate}

Any string that satisfies all these constraints is considered a correct solution, providing a robust "verifiable" accuracy metric for generative models that goes beyond simple multiple-choice selection or exact-match.

%% file: appendix/experiment_setup.tex
\section{Models \& Experimental Setup}
\label{appendix:models-experimental-setup}

\subsection{Model Access Names and Notes}
Below we list the canonical access names or placeholders used when calling each model, grouped by access method (Google, OpenAI, OpenRouter, Local). All models listed here are verified and accessible during the submission and release of this paper.

\paragraph{Google AI Studio}
\begin{itemize}[itemsep=0.1em, topsep=0.1em]
  \item \textbf{Gemini 3 Pro} \citep{google2025gemini3pro}: \texttt{gemini-3-pro-preview}
  \item \textbf{Gemini 2.5 Pro} \citep{comanici2025gemini}: \texttt{gemini-2.5-pro} % replace if different
\end{itemize}

\paragraph{OpenAI API}
\begin{itemize}[itemsep=0.1em, topsep=0.1em]
  \item \textbf{GPT-5} \citep{openai2025gpt5systemcard}: \texttt{gpt-5-2025-08-07}
  \item \textbf{o4-Mini} \citep{openai2025o4minisystemcard}: \texttt{o4-mini-2025-04-16}
\end{itemize}

\paragraph{OpenRouter}
\begin{itemize}[itemsep=0.1em, topsep=0.1em]
  \item \textbf{Claude Sonnet 4} \citep{anthropic2024claudesonnet4systemcard}: \texttt{anthropic/claude-sonnet-4}
  \item \textbf{Grok 4.1 Fast} \citep{xai2025grok41fast}: \texttt{x-ai/grok-4.1-fast}
  \item \textbf{Grok 4 Fast} \citep{xai2025grok4fast}: \texttt{x-ai/grok-4-fast}
  \item \textbf{GLM 4.6V} \citep{zai2025glm46v}: \texttt{z-ai/glm-4.6v\footnote{https://huggingface.co/zai-org/GLM-4.6V}}
  \item \textbf{GLM 4.5V} \citep{vteam2025glm45vglm41vthinkingversatilemultimodal}: \texttt{z-ai/glm-4.5v\footnote{https://huggingface.co/zai-org/GLM-4.5V}}
\end{itemize}

\paragraph{Local / vLLM}
\begin{itemize}[itemsep=0.1em, topsep=0.1em]
  \item \textbf{Qwen3-VL-235B} \citep{Qwen3-VL}: \texttt{Qwen/Qwen3-VL-235B-A22B-Thinking\footnote{https://huggingface.co/Qwen/Qwen3-VL-235B-A22B-Thinking}}
\end{itemize}

\subsection{vLLM Inference Settings}
For local vLLM runs, we used a server with 4x NVIDIA H100 80GB GPUs and CUDA 12.8, and ran the model with FP8 quantization. Generations used the model's default Qwen3 configuration; temperature and other sampling settings were chosen per the evaluation protocol below.

\subsection{Evaluation Protocol Details}
\begin{itemize}[itemsep=0.1em, topsep=0.1em]
  \item \textbf{Zero-shot:} All evaluations are zero-shot.
  \item \textbf{Temperature:} We used a temperature of 0 when possible to reduce sampling variability; when a model's API does not allow overriding temperature, we used the model's default.
  \item \textbf{Reasoning mode:} We used each model's default reasoning configuration except for Claude Sonnet 4 and Grok 4 Fast, for which we adjusted reasoning as described in the main text (see Section~\ref{sec:experiment-setup}).
  \item \textbf{Maximum output tokens:} For RAPM, we allowed the model's maximum permitted output tokens. For SWM and WCST, we limited per-turn outputs to 8192 tokens.
  \item \textbf{Thinking budget:} The system prompt described a thinking budget (not enforced as a hard token limit). For RAPM, we used an effective thinking budget of 30,000 tokens for models with a context limit of ~32K and 60,000 tokens for models with a context limit of 65K+. For SWM and WCST, the per-turn thinking budget was 4,000 tokens.
  \item \textbf{Repetitions:} SWM and WCST setups were run three times each to assess variability (data are generated on the fly); RAPM was run once per model.
  \item \textbf{WCST (notes):} For WCST runs that used note-taking assistance, we set the note window to 6.
\end{itemize}

\subsection{Reasoning Configuration}
Model reasoning capabilities were enabled through two complementary methods:

\paragraph{API Configuration.}
In our \texttt{ModelWrapper}, we programmatically enabled reasoning features where supported by the provider APIs.
\begin{itemize}[itemsep=0.1em, topsep=0.1em]
  \item \textbf{Google AI Studio:} We accessed Gemini models via the OpenAI-compatible endpoint and enabled reasoning by specifically configuring the \texttt{thinking\_config} within the \texttt{extra\_body} parameter. We set \texttt{include\_thoughts} to \texttt{True}, \texttt{thinking\_level} for Gemini 3 Pro (default to \texttt{high}), and \texttt{thinking\_budget} for Gemini 2.5 Pro (default to 8192).
  \item \textbf{OpenAI API:} For OpenAI models, reasoning is handled through \texttt{reasoning\_effort} parameter in the chat completion calls. We use the default  value \texttt{medium} for both GPT-5 and o4-Mini.
  \item \textbf{OpenRouter:} We set the \texttt{reasoning} parameter in the request body to \texttt{\{"exclude": False\}} to ensure reasoning tokens were generated and returned, and manage the thinking budget by setting \texttt{max\_tokens} within the reasoning configuration. For Grok, reasoning was explicitly enabled via \texttt{\{"enabled": True\}}.
  \item \textbf{vLLM (Local):} We enabled the \texttt{enable\_thinking} parameter in the chat template to activate reasoning generation.
\end{itemize}

\paragraph{Prompt-Based Instructions.}
For all tasks, we included explicit instructions in the prompt to guide the reasoning process, specifically when Chain-of-Thought (CoT) was enabled.

\begin{tcolorbox}[breakable, enhanced, top=1pt, left=1pt, right=1pt, bottom=1pt, title=Standardized CoT Prompt Addition]
  \small
  \textbf{SWM \& WCST:} \\
  Explain your thought process regarding the problem and the feedbacks you received in maximum \{think\_budget\} tokens wrapped with $<$think$>$ and $<$/think$>$. Then, provide a really short summary of your reasoning after the closing $<$/think$>$ tag.

  \textbf{RAPM:} \\
  Explain your thought process (max \{think\_budget\} tokens) inside $<$think$>$...$<$/think$>$ then give final answer.
\end{tcolorbox}

This dual approach ensured that models with native reasoning capabilities (activated via API) and standard models (guided via prompts) both engaged in structured reasoning before outputting a final answer.

\subsection{Disabling Reasoning}
\label{sec:disabling-reasoning}

We disable the reasoning capabilities for some models that are able to do so to conduct the ablation study (see Section~\ref{sec:reasoning-effect-rapm}) and to address specific model issues (specifically, Claude Sonnet 4 and Grok 4 Fast with ``overthinking'' that exhausted token limits on RAPM).

\paragraph{Configuration \& Prompts.}
For most models, reasoning was disabled by reversing the API configurations described previously (e.g., setting \texttt{include\_thoughts} to \texttt{False}, \texttt{reasoning} enabled to \texttt{False}, or \texttt{enable\_thinking} to \texttt{False}) and replacing the CoT prompt instructions with a standard direction to provide only the final answer, as shown below.

\begin{tcolorbox}[breakable, enhanced, top=1pt, left=1pt, right=1pt, bottom=1pt, title=Standardized Non-CoT Instruction]
  \small
  Answer only with your final answer.
\end{tcolorbox}

\paragraph{Gemini 2.5 Pro Specific Prompt.}
For Gemini 2.5 Pro, the model does not natively support disabling reasoning via API parameters. However, we found a way to enforce the no-reasoning condition by appending a specific critical instruction to override the model's default behavior.

\begin{tcolorbox}[breakable, enhanced, top=1pt, left=1pt, right=1pt, bottom=1pt, title=Gemini 2.5 Pro Non-CoT Instruction]
  \small
  CRITICAL INSTRUCTION: Do not use $<$thinking$>$ tags or reasoning steps. Provide the response immediately.
\end{tcolorbox}

%% file: appendix/benchmark_setup.tex
\section{External Benchmark Setup \& Results}
\label{sec:external-benchmark}

\paragraph{Long HotPotQA.} We work with the distractor development split of HotPotQA \citep{yang2018hotpotqadatasetdiverseexplainable}, sampling 500 question--passage pairs. To stress the models' turn tracking, we split each question and its supporting passages at the word level, treat every word as a separate turn, and fix the model's response on every intermediate turn to ``Okay.'' Each sequence is ordered with the full question first, followed by the concatenated passages. The results for Long HotPotQA are shown in Table \ref{tab:external-benchmarks}.

\paragraph{LLM Task Interference.} Following the official implementation from \citet{gupta2024llmtaskinterferenceinitial}, with minor updates, we evaluate two task/context pairings and sample 200 evaluation prompts each. The first pairing uses the MMLU AA task with Rotten Tomatoes exemplars as in-context data, while the second pairing is the Rotten Tomatoes task with TweetQA data. The results for LLM Task Interference are shown in Table \ref{tab:external-benchmarks}.

\paragraph{ARC-AGI 2.} Since the results for ARC-AGI 2 \citep{chollet2019measureintelligence} are already publicly available in their leaderboard website\footnote{https://arcprize.org/leaderboard}, we directly took the best results for each model, except for some models that are not available. Although the scores are very low, the results have been verified from the source. These are also summarized in Table \ref{tab:external-benchmarks}.

\input{appendix/external_benchmarks_table}

%% file: appendix/external_benchmarks_table.tex
\begin{table}[H]
\centering
\caption{Results for external benchmarks: Long HotPotQA, LLM Task Interference, and ARC-AGI 2. (-) indicates the model result is not available.}
\label{tab:external-benchmarks}
\begin{tabular}{lccc}
\toprule
\textbf{Model} & \textbf{Long HotPotQA} & \textbf{Task Interference} & \textbf{ARC-AGI 2} \\
\midrule
\midrule
GPT-5 & 0.936 & 0.958 & 0.099 \\
Gemini 3 Pro & 0.932 & 0.962 & 0.311 \\
Gemini 2.5 Pro & 0.924 & 0.875 & 0.049 \\
Claude Sonnet 4 & 0.930 & 0.948 & 0.059 \\
Grok 4.1 Fast & 0.910 & 0.913 & - \\
Grok 4 Fast & 0.461 & 0.788 & 0.053 \\
o4-Mini & 0.802 & 0.933 & 0.061 \\
GLM-4.6V & 0.862 & 0.938 & - \\
GLM-4.5V & 0.271 & 0.635 & - \\
Qwen3 VL 235B & 0.916 & 0.940 & - \\
\bottomrule
\end{tabular}
\end{table}

%% file: appendix/factor_analysis.tex
\section{Factor Analysis Details}
\label{appendix:factor-analysis}

Table~\ref{tab:factor_loadings} reports the factor loadings from the exploratory factor analysis described in Section~\ref{sec:factor-analysis-llm}.

\input{tables/factor_analysis_loadings}

%% file: tables/factor_analysis_loadings.tex
\begin{table}[H]
    \centering
    \caption{Factor analysis loadings for state-of-the-art LLM benchmarks. The single extracted factor (PA1) explains 75\% of the total variance, indicating a strong general capability factor ($g$) across diverse tasks.}
    \label{tab:factor_loadings}
    \begin{small} 
    \begin{tabularx}{0.65\linewidth}{Xc}
        \toprule
        \textbf{Benchmark} & \textbf{Loadings ($\lambda$)} \\
        \midrule
        \midrule
        GPQA Diamond \cite{rein2024gpqa} & 0.943 \\
        LiveCodeBench \cite{jainlivecodebench} & 0.926 \\
        SciCode \cite{tian2024scicode} & 0.922 \\
        AA-LCR \cite{artificialanalysis2025lcr} & 0.875 \\
        AIME 2025 \cite{aime25} & 0.870 \\
        IFBench \cite{pyatkin2025generalizing} & 0.861 \\
        Terminal Bench \cite{merrill2026terminal} & 0.834 \\
        MMLU Pro \cite{wang2024mmlu} & 0.825 \\
        $\tau$-Bench \cite{barres2025tau2benchevaluatingconversationalagents} & 0.818 \\
        Humanity's Last Exam \cite{phan2025humanity} & 0.768 \\
        \bottomrule
    \end{tabularx}
    \end{small}
\end{table}

%% file: appendix/full_results.tex
\newpage
\section{Full Results}
\label{sec:full-results}
% We provide complete results and some sample qualitative samples for all evaluated models for RAPM, SWM and WCST tests.

\subsection{Raven's Advanced Progressive Matrices (RAPM)}
The complete results for RAPM image tests, broken down by pattern type, are provided in Table \ref{tab:rapm-full-results}. We also provide qualitative examples of reasoning failures across text and image modalities in Table \ref{tab:rapm-failure-examples}, highlighting the prevalence of hallucination and over-interpretation.
\input{appendix/rapm_full_results}
\input{appendix/rapm_failure_examples}

\subsection{Spatial Working Memory (SWM)}
The complete results for SWM tests are provided in Table \ref{tab:swm_text} for text modality, in Table \ref{tab:swm_image} for image modality, and in Table~\ref{tab:swm_image-text} for image-text modality. All metrics are averaged over three runs.
\input{appendix/swm_text_full_results}
\input{appendix/swm_image_full_results}
\input{appendix/swm_image_text_full_results}
% The SWM failure breakdown by error type (also shown as Figure~\ref{fig:swm-wcst-failures} in the main paper) is provided in Table~\ref{tab:swm-failures} for reference.
% \input{tables/failure_swm_table}

\clearpage
\subsection{Wisconsin Card Sorting Test (WCST)}
The complete results for WCST tests are provided in Table \ref{tab:wcst_text} for text modality, and in Table \ref{tab:wcst_image} for image modality. All metrics are averaged over three runs.
\input{appendix/wcst_full_results}
% The WCST FMS and PR error rates (also shown as Figure~\ref{fig:swm-wcst-failures} in the main paper) are provided in Table~\ref{tab:wcst-failures} for reference.
% \input{tables/failure_wcst_table}

%% file: appendix/rapm_full_results.tex
% RAPM detailed results
\begin{table}[H]
\centering
\small
\caption{RAPM image accuracy breakdown by pattern type.}
\resizebox{\textwidth}{!}{%
\begin{tabular}{lrrrrrrr}
\toprule
\textbf{Model} & \textbf{Dist. Four} & \textbf{Dist. Nine} & \textbf{Center} & \textbf{In-Dist. Out-Center} & \textbf{In-Out Center} & \textbf{Left-Right} & \textbf{Up-Down} \\
\midrule\midrule
GPT-5 & 0.40 & 0.35 & 0.25 & 0.15 & 0.35 & 0.05 & 0.40 \\
Gemini 3 Pro & 0.55 & 0.65 & 0.55 & 0.35 & 0.20 & 0.25 & 0.25 \\
Gemini 2.5 Pro & 0.05 & 0.10 & 0.35 & 0.20 & 0.40 & 0.20 & 0.25 \\
Claude Sonnet 4 & 0.30 & 0.45 & 0.15 & 0.20 & 0.20 & 0.10 & 0.15 \\
Grok 4.1 Fast & 0.15 & 0.20 & 0.15 & 0.05 & 0.00 & 0.00 & 0.15 \\
Grok 4 Fast & 0.05 & 0.10 & 0.05 & 0.25 & 0.10 & 0.10 & 0.05 \\
o4-Mini & 0.45 & 0.15 & 0.25 & 0.10 & 0.30 & 0.05 & 0.15 \\
GLM 4.6V & 0.45 & 0.30 & 0.40 & 0.35 & 0.30 & 0.25 & 0.45 \\
GLM 4.5V & 0.20 & 0.05 & 0.15 & 0.10 & 0.15 & 0.20 & 0.05 \\
Qwen3-VL-235B & 0.45 & 0.65 & 0.55 & 0.40 & 0.40 & 0.50 & 0.50 \\
\bottomrule
\end{tabular}
}
\label{tab:rapm-full-results}
\end{table}

%% file: appendix/rapm_failure_examples.tex
% RAPM Failure Examples
\begin{table}[H]
  \centering
  \small
  \caption{Selected reasoning failure examples in RAPM (image and text). Models frequently hallucinate attributes to fit a complex hypothesis ("visual/numerological Hallucination") or prefer abstract semantic rules over simple local operations ("semantic over-interpretation").}
  \label{tab:rapm-failure-examples}
  \resizebox{\textwidth}{!}{%
    \begin{tabular}{p{0.12\textwidth} p{0.12\textwidth} p{0.14\textwidth} p{0.38\textwidth} p{0.16\textwidth}}
      \toprule
      \textbf{Task/Model} & \textbf{ID} & \textbf{Pattern/Context} & \textbf{Reasoning Snippet (Failure Point)} & \textbf{Outcome} \\
      \midrule
      \midrule
      \textbf{Image} \newline GPT-5 & RAVEN\_238 \newline center & Up-Down \newline (Split layout) & "...Row 3: top becomes \textbf{large solid black triangle}. Bottom shapes by row show a sides-count sequence decreasing... 5$\to$4$\to$3..." \newline \textcolor{red}{\textit{(Visual Hallucination: No such triangle exists.)}} & Pred: 2 \newline Correct: 0 \\
      \midrule
      \textbf{Image} \newline GPT-5 & RAVEN\_238 \newline dist\_nine & Dist. Nine \newline (3x3 Grid logic) & "...Row 1: Left has all 9 positions filled... middle emphasizes corners... The combination is still all 9 positions..." \newline \textcolor{red}{\textit{(Relational Over-complexity: Hallucinates a global superposition rule for a simple local pattern.)}} & Pred: 5 \newline Correct: 2 \\
      \midrule
      \textbf{Text MC} \newline Gemini 3 Pro & text\_rapm\_0 & Simple Arithmetic \newline (Char length) & "Row 1: 6, 10, 13 characters... Row 2: 6, 9, 12... notice a potential arithmetic-like progression" \newline \textcolor{red}{\textit{(Numerological Hallucination: Forces arithmetic series on string lengths where rule is character types.)}} & Pred: 3 \newline Correct: 5 \\
      \midrule
      \textbf{Text MC} \newline GPT-5 & text\_rapm\_4 & Column Count \newline (Uppercase) & "...Column 2 uppercase counts go 0... 1... 2. So for Column 3 they should be 0... 1... 2." \newline \textcolor{red}{\textit{(False Generalization: Hallucinates column-wise progression that doesn't exist in data.)}} & Pred: 3 \newline Correct: 6 \\
      \midrule
      \textbf{Text Gen} \newline GPT-5 & text\_rapm\_2 & Constraint Satisfaction \newline (Vowel Count) & "...Rows keep a constant number of uppercase letters... Column 3 needs 6 vowels... A valid completion is 'aeiOuEsTH'." \newline \textcolor{red}{\textit{(Metadata vs Content: Hallucinates content string that fits metadata counts but fails derivation logic.)}} & Pred: aeiOuEsTH \newline Rules: 5 lowercase vowels, 4 uppercase letters \\
      \midrule
    \textbf{Text Gen} \newline Gemini 3 Pro & text\_rapm\_3 & String Length \newline (Decreasing) & "Column 2 seems to be decreasing length... Row 3: 6, 9, ?... Missing string should be length 8." \newline \textcolor{red}{\textit{(Numerological: Obsesses over length counts, ignoring character content patterns.)}} & Pred: T4Z]LBEBUM6 \newline Rules: Col progression of uppercase vowels (2$\to$3$\to$4) \\
    \bottomrule
  \end{tabular}
}
\end{table}

%% file: appendix/swm_text_full_results.tex
\begin{table}[H]
  \centering
  \small
  \caption{Spatial Working Memory full results (text modality). All metrics are reported as mean with standard deviation (std) in a subscript.}
  \resizebox{\textwidth}{!}{%
    \begin{tabular}{ll*{7}{l}}
      \toprule
      \multicolumn{1}{c}{\textbf{Difficulty}} & \multicolumn{1}{c}{\textbf{Model}} & \multicolumn{1}{c}{$\boldsymbol{S}_{swm}$} & \multicolumn{1}{c}{\textbf{Tokens Score}} & \multicolumn{1}{c}{\textbf{Guesses}} & \multicolumn{1}{c}{\textbf{Illegal}} & \multicolumn{1}{c}{\textbf{Invalid}} & \multicolumn{1}{c}{\textbf{Repeated}} & \multicolumn{1}{c}{\textbf{No-box}} \\
      \midrule
      \midrule
      \multirow{10}{*}{Easy}
      & GPT-5 & $1.000_{\scriptscriptstyle \pm .00}$ & $1.000_{\scriptscriptstyle \pm .00}$ & $18.000_{\scriptscriptstyle \pm 6.16}$ & $0.000_{\scriptscriptstyle \pm 0.00}$ & $0.000_{\scriptscriptstyle \pm 0.00}$ & $0.000_{\scriptscriptstyle \pm 0.00}$ & $0.000_{\scriptscriptstyle \pm 0.00}$ \\
      & Gemini 3 Pro & $1.000_{\scriptscriptstyle \pm .00}$ & $1.000_{\scriptscriptstyle \pm .00}$ & $18.000_{\scriptscriptstyle \pm 2.94}$ & $0.000_{\scriptscriptstyle \pm 0.00}$ & $0.000_{\scriptscriptstyle \pm 0.00}$ & $0.000_{\scriptscriptstyle \pm 0.00}$ & $0.000_{\scriptscriptstyle \pm 0.00}$ \\
      & Gemini 2.5 Pro & $1.000_{\scriptscriptstyle \pm .00}$ & $1.000_{\scriptscriptstyle \pm .00}$ & $19.667_{\scriptscriptstyle \pm 1.70}$ & $0.000_{\scriptscriptstyle \pm 0.00}$ & $0.000_{\scriptscriptstyle \pm 0.00}$ & $0.000_{\scriptscriptstyle \pm 0.00}$ & $0.000_{\scriptscriptstyle \pm 0.00}$ \\
      & Claude Sonnet 4 & $1.000_{\scriptscriptstyle \pm .00}$ & $1.000_{\scriptscriptstyle \pm .00}$ & $23.000_{\scriptscriptstyle \pm 3.56}$ & $0.000_{\scriptscriptstyle \pm 0.00}$ & $0.000_{\scriptscriptstyle \pm 0.00}$ & $0.000_{\scriptscriptstyle \pm 0.00}$ & $0.000_{\scriptscriptstyle \pm 0.00}$ \\
      & Grok 4.1 Fast & $1.000_{\scriptscriptstyle \pm .00}$ & $1.000_{\scriptscriptstyle \pm .00}$ & $26.667_{\scriptscriptstyle \pm 2.87}$ & $0.000_{\scriptscriptstyle \pm 0.00}$ & $0.000_{\scriptscriptstyle \pm 0.00}$ & $0.000_{\scriptscriptstyle \pm 0.00}$ & $0.000_{\scriptscriptstyle \pm 0.00}$ \\
      & Grok 4 Fast & $1.000_{\scriptscriptstyle \pm .00}$ & $1.000_{\scriptscriptstyle \pm .00}$ & $18.333_{\scriptscriptstyle \pm 3.40}$ & $0.000_{\scriptscriptstyle \pm 0.00}$ & $0.333_{\scriptscriptstyle \pm 0.47}$ & $0.000_{\scriptscriptstyle \pm 0.00}$ & $0.000_{\scriptscriptstyle \pm 0.00}$ \\
      & o4-Mini & $0.951_{\scriptscriptstyle \pm .05}$ & $1.000_{\scriptscriptstyle \pm .00}$ & $23.667_{\scriptscriptstyle \pm 4.03}$ & $0.000_{\scriptscriptstyle \pm 0.00}$ & $0.667_{\scriptscriptstyle \pm 0.94}$ & $1.000_{\scriptscriptstyle \pm 0.82}$ & $0.000_{\scriptscriptstyle \pm 0.00}$ \\
      & GLM 4.6V & $1.000_{\scriptscriptstyle \pm .00}$ & $1.000_{\scriptscriptstyle \pm .00}$ & $24.000_{\scriptscriptstyle \pm 2.94}$ & $0.000_{\scriptscriptstyle \pm 0.00}$ & $3.333_{\scriptscriptstyle \pm 0.94}$ & $0.000_{\scriptscriptstyle \pm 0.00}$ & $0.000_{\scriptscriptstyle \pm 0.00}$ \\
      & GLM 4.5V & $1.000_{\scriptscriptstyle \pm .00}$ & $1.000_{\scriptscriptstyle \pm .00}$ & $20.333_{\scriptscriptstyle \pm 1.70}$ & $0.000_{\scriptscriptstyle \pm 0.00}$ & $0.000_{\scriptscriptstyle \pm 0.00}$ & $0.000_{\scriptscriptstyle \pm 0.00}$ & $0.000_{\scriptscriptstyle \pm 0.00}$ \\
      & Qwen3-VL-235B & $1.000_{\scriptscriptstyle \pm .00}$ & $1.000_{\scriptscriptstyle \pm .00}$ & $20.000_{\scriptscriptstyle \pm 2.45}$ & $0.000_{\scriptscriptstyle \pm 0.00}$ & $0.000_{\scriptscriptstyle \pm 0.00}$ & $0.000_{\scriptscriptstyle \pm 0.00}$ & $0.000_{\scriptscriptstyle \pm 0.00}$ \\
      \midrule
      \multirow{10}{*}{Hard}
      & GPT-5 & $1.000_{\scriptscriptstyle \pm .00}$ & $1.000_{\scriptscriptstyle \pm .00}$ & $56.333_{\scriptscriptstyle \pm 1.25}$ & $0.000_{\scriptscriptstyle \pm 0.00}$ & $0.000_{\scriptscriptstyle \pm 0.00}$ & $0.000_{\scriptscriptstyle \pm 0.00}$ & $0.000_{\scriptscriptstyle \pm 0.00}$ \\
      & Gemini 3 Pro & $1.000_{\scriptscriptstyle \pm .00}$ & $1.000_{\scriptscriptstyle \pm .00}$ & $63.333_{\scriptscriptstyle \pm 1.70}$ & $0.000_{\scriptscriptstyle \pm 0.00}$ & $1.000_{\scriptscriptstyle \pm 0.82}$ & $0.000_{\scriptscriptstyle \pm 0.00}$ & $0.000_{\scriptscriptstyle \pm 0.00}$ \\
      & Gemini 2.5 Pro & $0.980_{\scriptscriptstyle \pm .03}$ & $1.000_{\scriptscriptstyle \pm .00}$ & $82.333_{\scriptscriptstyle \pm 13.57}$ & $0.000_{\scriptscriptstyle \pm 0.00}$ & $0.000_{\scriptscriptstyle \pm 0.00}$ & $2.000_{\scriptscriptstyle \pm 2.83}$ & $0.000_{\scriptscriptstyle \pm 0.00}$ \\
      & Claude Sonnet 4 & $0.866_{\scriptscriptstyle \pm .18}$ & $1.000_{\scriptscriptstyle \pm .00}$ & $78.333_{\scriptscriptstyle \pm 11.44}$ & $0.000_{\scriptscriptstyle \pm 0.00}$ & $0.333_{\scriptscriptstyle \pm 0.47}$ & $12.333_{\scriptscriptstyle \pm 16.74}$ & $0.000_{\scriptscriptstyle \pm 0.00}$ \\
      & Grok 4.1 Fast & $0.983_{\scriptscriptstyle \pm .01}$ & $1.000_{\scriptscriptstyle \pm .00}$ & $52.667_{\scriptscriptstyle \pm 9.29}$ & $0.000_{\scriptscriptstyle \pm 0.00}$ & $0.000_{\scriptscriptstyle \pm 0.00}$ & $1.000_{\scriptscriptstyle \pm 0.82}$ & $0.000_{\scriptscriptstyle \pm 0.00}$ \\
      & Grok 4 Fast & $0.994_{\scriptscriptstyle \pm .01}$ & $1.000_{\scriptscriptstyle \pm .00}$ & $63.000_{\scriptscriptstyle \pm 4.55}$ & $0.333_{\scriptscriptstyle \pm 0.47}$ & $0.333_{\scriptscriptstyle \pm 0.47}$ & $0.000_{\scriptscriptstyle \pm 0.00}$ & $0.000_{\scriptscriptstyle \pm 0.00}$ \\
      & o4-Mini & $0.811_{\scriptscriptstyle \pm .07}$ & $1.000_{\scriptscriptstyle \pm .00}$ & $79.333_{\scriptscriptstyle \pm 1.70}$ & $5.333_{\scriptscriptstyle \pm 3.77}$ & $0.000_{\scriptscriptstyle \pm 0.00}$ & $9.667_{\scriptscriptstyle \pm 2.36}$ & $0.000_{\scriptscriptstyle \pm 0.00}$ \\
      & GLM 4.6V & $0.574_{\scriptscriptstyle \pm .35}$ & $0.694_{\scriptscriptstyle \pm .27}$ & $129.333_{\scriptscriptstyle \pm 20.74}$ & $3.000_{\scriptscriptstyle \pm 3.56}$ & $33.333_{\scriptscriptstyle \pm 34.50}$ & $32.333_{\scriptscriptstyle \pm 44.32}$ & $0.000_{\scriptscriptstyle \pm 0.00}$ \\
      & GLM 4.5V & $0.536_{\scriptscriptstyle \pm .24}$ & $0.736_{\scriptscriptstyle \pm .37}$ & $105.333_{\scriptscriptstyle \pm 27.64}$ & $16.667_{\scriptscriptstyle \pm 11.81}$ & $42.667_{\scriptscriptstyle \pm 58.93}$ & $0.333_{\scriptscriptstyle \pm 0.47}$ & $0.000_{\scriptscriptstyle \pm 0.00}$ \\
      & Qwen3-VL-235B & $0.760_{\scriptscriptstyle \pm .06}$ & $1.000_{\scriptscriptstyle \pm .00}$ & $80.333_{\scriptscriptstyle \pm 10.66}$ & $19.333_{\scriptscriptstyle \pm 7.72}$ & $1.000_{\scriptscriptstyle \pm 0.00}$ & $0.333_{\scriptscriptstyle \pm 0.47}$ & $0.000_{\scriptscriptstyle \pm 0.00}$ \\
      \midrule
      \multirow{8}{*}{\shortstack[l]{Hard - Notes}}
      & Gemini 2.5 Pro & $0.965_{\scriptscriptstyle \pm .04}$ & $1.000_{\scriptscriptstyle \pm .00}$ & $66.667_{\scriptscriptstyle \pm 9.18}$ & $0.000_{\scriptscriptstyle \pm 0.00}$ & $0.000_{\scriptscriptstyle \pm 0.00}$ & $2.333_{\scriptscriptstyle \pm 2.62}$ & $0.000_{\scriptscriptstyle \pm 0.00}$ \\
      & Claude Sonnet 4 & $0.826_{\scriptscriptstyle \pm .06}$ & $1.000_{\scriptscriptstyle \pm .00}$ & $81.000_{\scriptscriptstyle \pm 9.27}$ & $1.000_{\scriptscriptstyle \pm 0.82}$ & $0.000_{\scriptscriptstyle \pm 0.00}$ & $13.667_{\scriptscriptstyle \pm 5.56}$ & $0.000_{\scriptscriptstyle \pm 0.00}$ \\
      & Grok 4.1 Fast & $0.973_{\scriptscriptstyle \pm .01}$ & $1.000_{\scriptscriptstyle \pm .00}$ & $62.333_{\scriptscriptstyle \pm 3.30}$ & $0.333_{\scriptscriptstyle \pm 0.47}$ & $0.000_{\scriptscriptstyle \pm 0.00}$ & $1.333_{\scriptscriptstyle \pm 0.47}$ & $0.000_{\scriptscriptstyle \pm 0.00}$ \\
      & Grok 4 Fast & $0.964_{\scriptscriptstyle \pm .01}$ & $1.000_{\scriptscriptstyle \pm .00}$ & $65.333_{\scriptscriptstyle \pm 1.25}$ & $0.333_{\scriptscriptstyle \pm 0.47}$ & $0.000_{\scriptscriptstyle \pm 0.00}$ & $2.000_{\scriptscriptstyle \pm 0.00}$ & $0.000_{\scriptscriptstyle \pm 0.00}$ \\
      & o4-Mini & $0.935_{\scriptscriptstyle \pm .03}$ & $1.000_{\scriptscriptstyle \pm .00}$ & $75.333_{\scriptscriptstyle \pm 15.92}$ & $1.000_{\scriptscriptstyle \pm 1.41}$ & $0.333_{\scriptscriptstyle \pm 0.47}$ & $4.333_{\scriptscriptstyle \pm 2.49}$ & $0.000_{\scriptscriptstyle \pm 0.00}$ \\
      & GLM 4.6V & $0.803_{\scriptscriptstyle \pm .14}$ & $1.000_{\scriptscriptstyle \pm .00}$ & $85.000_{\scriptscriptstyle \pm 20.22}$ & $0.333_{\scriptscriptstyle \pm 0.47}$ & $1.000_{\scriptscriptstyle \pm 0.00}$ & $19.000_{\scriptscriptstyle \pm 16.87}$ & $0.000_{\scriptscriptstyle \pm 0.00}$ \\
      & GLM 4.5V & $0.851_{\scriptscriptstyle \pm .09}$ & $1.000_{\scriptscriptstyle \pm .00}$ & $79.333_{\scriptscriptstyle \pm 11.12}$ & $10.333_{\scriptscriptstyle \pm 7.36}$ & $14.333_{\scriptscriptstyle \pm 16.11}$ & $0.333_{\scriptscriptstyle \pm 0.47}$ & $0.000_{\scriptscriptstyle \pm 0.00}$ \\
      & Qwen3-VL-235B & $0.782_{\scriptscriptstyle \pm .23}$ & $0.972_{\scriptscriptstyle \pm .04}$ & $92.333_{\scriptscriptstyle \pm 36.81}$ & $0.667_{\scriptscriptstyle \pm 0.94}$ & $2.333_{\scriptscriptstyle \pm 3.30}$ & $25.667_{\scriptscriptstyle \pm 32.83}$ & $0.000_{\scriptscriptstyle \pm 0.00}$ \\
      \bottomrule
    \end{tabular}%
  }
  \label{tab:swm_text}
\end{table}

%% file: appendix/swm_image_full_results.tex
\begin{table}[H]
    \centering
    \small
    \caption{Spatial Working Memory full results (image modality). All metrics are reported as mean with standard deviation (std) in a subscript.}
    \resizebox{\textwidth}{!}{%
    \begin{tabular}{ll*{7}{l}}
        \toprule
        \multicolumn{1}{c}{\textbf{Difficulty}} & \multicolumn{1}{c}{\textbf{Model}} & \multicolumn{1}{c}{$\boldsymbol{S}_{swm}$} & \multicolumn{1}{c}{\textbf{Tokens Score}} & \multicolumn{1}{c}{\textbf{Guesses}} & \multicolumn{1}{c}{\textbf{Illegal}} & \multicolumn{1}{c}{\textbf{Invalid}} & \multicolumn{1}{c}{\textbf{Repeated}} & \multicolumn{1}{c}{\textbf{No-box}} \\
        \midrule
        \midrule
        \multirow{10}{*}{Easy}
        & GPT-5 & $0.927_{\scriptscriptstyle \pm 0.08}$ & $1.000_{\scriptscriptstyle \pm 0.00}$ & $27.667_{\scriptscriptstyle \pm 4.64}$ & $0.000_{\scriptscriptstyle \pm 0.00}$ & $0.000_{\scriptscriptstyle \pm 0.00}$ & $2.000_{\scriptscriptstyle \pm 2.83}$ & $0.333_{\scriptscriptstyle \pm 0.47}$ \\
        & Gemini 3 Pro & $1.000_{\scriptscriptstyle \pm 0.00}$ & $1.000_{\scriptscriptstyle \pm 0.00}$ & $22.333_{\scriptscriptstyle \pm 0.94}$ & $0.000_{\scriptscriptstyle \pm 0.00}$ & $0.000_{\scriptscriptstyle \pm 0.00}$ & $0.000_{\scriptscriptstyle \pm 0.00}$ & $0.000_{\scriptscriptstyle \pm 0.00}$ \\
        & Gemini 2.5 Pro & $0.408_{\scriptscriptstyle \pm 0.38}$ & $0.625_{\scriptscriptstyle \pm 0.44}$ & $51.333_{\scriptscriptstyle \pm 17.91}$ & $0.000_{\scriptscriptstyle \pm 0.00}$ & $0.000_{\scriptscriptstyle \pm 0.00}$ & $2.667_{\scriptscriptstyle \pm 3.77}$ & $30.667_{\scriptscriptstyle \pm 20.42}$ \\
        & Claude Sonnet 4 & $0.630_{\scriptscriptstyle \pm 0.41}$ & $0.750_{\scriptscriptstyle \pm 0.35}$ & $43.667_{\scriptscriptstyle \pm 14.43}$ & $0.000_{\scriptscriptstyle \pm 0.00}$ & $0.000_{\scriptscriptstyle \pm 0.00}$ & $4.000_{\scriptscriptstyle \pm 5.66}$ & $14.000_{\scriptscriptstyle \pm 15.75}$ \\
        & Grok 4.1 Fast & $0.014_{\scriptscriptstyle \pm 0.01}$ & $0.125_{\scriptscriptstyle \pm 0.10}$ & $64.000_{\scriptscriptstyle \pm 0.00}$ & $0.333_{\scriptscriptstyle \pm 0.47}$ & $0.000_{\scriptscriptstyle \pm 0.00}$ & $5.000_{\scriptscriptstyle \pm 5.72}$ & $51.333_{\scriptscriptstyle \pm 5.25}$ \\
        & Grok 4 Fast & $0.029_{\scriptscriptstyle \pm 0.04}$ & $0.125_{\scriptscriptstyle \pm 0.18}$ & $64.000_{\scriptscriptstyle \pm 0.00}$ & $0.000_{\scriptscriptstyle \pm 0.00}$ & $0.333_{\scriptscriptstyle \pm 0.47}$ & $14.333_{\scriptscriptstyle \pm 13.27}$ & $42.000_{\scriptscriptstyle \pm 10.61}$ \\
        & o4-Mini & $0.042_{\scriptscriptstyle \pm 0.04}$ & $0.208_{\scriptscriptstyle \pm 0.21}$ & $64.000_{\scriptscriptstyle \pm 0.00}$ & $0.000_{\scriptscriptstyle \pm 0.00}$ & $1.333_{\scriptscriptstyle \pm 0.47}$ & $30.667_{\scriptscriptstyle \pm 3.09}$ & $21.333_{\scriptscriptstyle \pm 0.94}$ \\
        & GLM 4.6V & $0.000_{\scriptscriptstyle \pm 0.00}$ & $0.000_{\scriptscriptstyle \pm 0.00}$ & $64.000_{\scriptscriptstyle \pm 0.00}$ & $0.000_{\scriptscriptstyle \pm 0.00}$ & $63.333_{\scriptscriptstyle \pm 0.47}$ & $0.000_{\scriptscriptstyle \pm 0.00}$ & $0.000_{\scriptscriptstyle \pm 0.00}$ \\
        & GLM 4.5V & $0.000_{\scriptscriptstyle \pm 0.00}$ & $0.000_{\scriptscriptstyle \pm 0.00}$ & $64.000_{\scriptscriptstyle \pm 0.00}$ & $0.000_{\scriptscriptstyle \pm 0.00}$ & $32.000_{\scriptscriptstyle \pm 2.83}$ & $6.333_{\scriptscriptstyle \pm 6.34}$ & $23.333_{\scriptscriptstyle \pm 5.25}$ \\
        & Qwen3-VL-235B & $0.031_{\scriptscriptstyle \pm 0.02}$ & $0.208_{\scriptscriptstyle \pm 0.12}$ & $64.000_{\scriptscriptstyle \pm 0.00}$ & $12.667_{\scriptscriptstyle \pm 11.09}$ & $2.667_{\scriptscriptstyle \pm 3.77}$ & $26.333_{\scriptscriptstyle \pm 20.34}$ & $14.000_{\scriptscriptstyle \pm 13.37}$ \\
        \midrule
        \multirow{10}{*}{Hard}
        & GPT-5 & $0.631_{\scriptscriptstyle \pm 0.16}$ & $1.000_{\scriptscriptstyle \pm 0.00}$ & $98.667_{\scriptscriptstyle \pm 19.07}$ & $10.333_{\scriptscriptstyle \pm 2.49}$ & $2.000_{\scriptscriptstyle \pm 2.83}$ & $14.667_{\scriptscriptstyle \pm 12.26}$ & $13.333_{\scriptscriptstyle \pm 18.15}$ \\
        & Gemini 3 Pro & $0.793_{\scriptscriptstyle \pm 0.24}$ & $0.958_{\scriptscriptstyle \pm 0.06}$ & $90.000_{\scriptscriptstyle \pm 38.88}$ & $20.000_{\scriptscriptstyle \pm 27.58}$ & $1.000_{\scriptscriptstyle \pm 0.82}$ & $4.667_{\scriptscriptstyle \pm 4.11}$ & $0.000_{\scriptscriptstyle \pm 0.00}$ \\
        & Gemini 2.5 Pro & $0.622_{\scriptscriptstyle \pm 0.30}$ & $0.944_{\scriptscriptstyle \pm 0.08}$ & $97.333_{\scriptscriptstyle \pm 34.92}$ & $35.000_{\scriptscriptstyle \pm 30.82}$ & $1.000_{\scriptscriptstyle \pm 1.41}$ & $9.000_{\scriptscriptstyle \pm 11.34}$ & $0.333_{\scriptscriptstyle \pm 0.47}$ \\
        & Claude Sonnet 4 & $0.005_{\scriptscriptstyle \pm 0.00}$ & $0.056_{\scriptscriptstyle \pm 0.02}$ & $144.000_{\scriptscriptstyle \pm 0.00}$ & $0.000_{\scriptscriptstyle \pm 0.00}$ & $44.333_{\scriptscriptstyle \pm 0.47}$ & $8.000_{\scriptscriptstyle \pm 3.74}$ & $82.667_{\scriptscriptstyle \pm 2.87}$ \\
        & Grok 4.1 Fast & $0.009_{\scriptscriptstyle \pm 0.01}$ & $0.069_{\scriptscriptstyle \pm 0.04}$ & $144.000_{\scriptscriptstyle \pm 0.00}$ & $0.000_{\scriptscriptstyle \pm 0.00}$ & $25.333_{\scriptscriptstyle \pm 26.39}$ & $7.333_{\scriptscriptstyle \pm 8.99}$ & $99.667_{\scriptscriptstyle \pm 27.98}$ \\
        & Grok 4 Fast & $0.005_{\scriptscriptstyle \pm 0.01}$ & $0.056_{\scriptscriptstyle \pm 0.05}$ & $144.000_{\scriptscriptstyle \pm 0.00}$ & $0.000_{\scriptscriptstyle \pm 0.00}$ & $39.000_{\scriptscriptstyle \pm 53.03}$ & $7.667_{\scriptscriptstyle \pm 7.04}$ & $88.333_{\scriptscriptstyle \pm 42.76}$ \\
        & o4-Mini & $0.005_{\scriptscriptstyle \pm 0.00}$ & $0.056_{\scriptscriptstyle \pm 0.02}$ & $144.000_{\scriptscriptstyle \pm 0.00}$ & $0.000_{\scriptscriptstyle \pm 0.00}$ & $1.333_{\scriptscriptstyle \pm 1.25}$ & $32.000_{\scriptscriptstyle \pm 13.49}$ & $98.333_{\scriptscriptstyle \pm 13.47}$ \\
        & GLM 4.6V & $0.028_{\scriptscriptstyle \pm 0.04}$ & $0.069_{\scriptscriptstyle \pm 0.10}$ & $144.000_{\scriptscriptstyle \pm 0.00}$ & $0.000_{\scriptscriptstyle \pm 0.00}$ & $119.333_{\scriptscriptstyle \pm 34.88}$ & $8.333_{\scriptscriptstyle \pm 11.79}$ & $6.333_{\scriptscriptstyle \pm 8.96}$ \\
        & GLM 4.5V & $0.000_{\scriptscriptstyle \pm 0.00}$ & $0.000_{\scriptscriptstyle \pm 0.00}$ & $144.000_{\scriptscriptstyle \pm 0.00}$ & $0.000_{\scriptscriptstyle \pm 0.00}$ & $111.667_{\scriptscriptstyle \pm 0.94}$ & $0.667_{\scriptscriptstyle \pm 0.94}$ & $30.667_{\scriptscriptstyle \pm 2.05}$ \\
        & Qwen3-VL-235B & $0.135_{\scriptscriptstyle \pm 0.12}$ & $0.347_{\scriptscriptstyle \pm 0.27}$ & $144.000_{\scriptscriptstyle \pm 0.00}$ & $4.333_{\scriptscriptstyle \pm 6.13}$ & $63.000_{\scriptscriptstyle \pm 1.41}$ & $30.667_{\scriptscriptstyle \pm 14.06}$ & $21.000_{\scriptscriptstyle \pm 19.87}$ \\
        \bottomrule
    \end{tabular}%
    }
    \label{tab:swm_image}
\end{table}

%% file: appendix/swm_image_text_full_results.tex
\begin{table}[H]
  \centering
  \small
  \caption{Spatial Working Memory full results (image-text modality). All metrics are reported as mean with standard deviation (std) in a subscript.}
  \resizebox{\textwidth}{!}{%
    \begin{tabular}{ll*{7}{l}}
      \toprule
      \multicolumn{1}{c}{\textbf{Difficulty}} & \multicolumn{1}{c}{\textbf{Model}} & \multicolumn{1}{c}{$\boldsymbol{S}_{swm}$} & \multicolumn{1}{c}{\textbf{Tokens Score}} & \multicolumn{1}{c}{\textbf{Guesses}} & \multicolumn{1}{c}{\textbf{Illegal}} & \multicolumn{1}{c}{\textbf{Invalid}} & \multicolumn{1}{c}{\textbf{Repeated}} & \multicolumn{1}{c}{\textbf{No-box}} \\
      \midrule
      \midrule
      \multirow{10}{*}{Easy}
      & GPT-5 & $0.986_{\scriptscriptstyle \pm .02}$ & $1.000_{\scriptscriptstyle \pm .00}$ & $24.000_{\scriptscriptstyle \pm 2.16}$ & $0.000_{\scriptscriptstyle \pm 0.00}$ & $0.000_{\scriptscriptstyle \pm 0.00}$ & $0.333_{\scriptscriptstyle \pm 0.47}$ & $0.000_{\scriptscriptstyle \pm 0.00}$ \\
      & Gemini 3 Pro & $1.000_{\scriptscriptstyle \pm .00}$ & $1.000_{\scriptscriptstyle \pm .00}$ & $20.667_{\scriptscriptstyle \pm 1.70}$ & $0.000_{\scriptscriptstyle \pm 0.00}$ & $0.000_{\scriptscriptstyle \pm 0.00}$ & $0.000_{\scriptscriptstyle \pm 0.00}$ & $0.000_{\scriptscriptstyle \pm 0.00}$ \\
      & Gemini 2.5 Pro & $0.733_{\scriptscriptstyle \pm .28}$ & $0.958_{\scriptscriptstyle \pm 0.06}$ & $36.667_{\scriptscriptstyle \pm 19.60}$ & $0.000_{\scriptscriptstyle \pm 0.00}$ & $0.000_{\scriptscriptstyle \pm 0.00}$ & $0.000_{\scriptscriptstyle \pm 0.00}$ & $14.000_{\scriptscriptstyle \pm 17.68}$ \\
      & Claude Sonnet 4 & $0.256_{\scriptscriptstyle \pm .18}$ & $0.625_{\scriptscriptstyle \pm 0.37}$ & $59.333_{\scriptscriptstyle \pm 6.60}$ & $0.000_{\scriptscriptstyle \pm 0.00}$ & $0.000_{\scriptscriptstyle \pm 0.00}$ & $1.667_{\scriptscriptstyle \pm 2.36}$ & $39.000_{\scriptscriptstyle \pm 12.36}$ \\
      & Grok 4.1 Fast & $0.145_{\scriptscriptstyle \pm .20}$ & $0.375_{\scriptscriptstyle \pm 0.44}$ & $53.667_{\scriptscriptstyle \pm 14.61}$ & $0.000_{\scriptscriptstyle \pm 0.00}$ & $0.667_{\scriptscriptstyle \pm 0.94}$ & $4.333_{\scriptscriptstyle \pm 5.44}$ & $40.333_{\scriptscriptstyle \pm 16.94}$ \\
      & Grok 4 Fast & $0.021_{\scriptscriptstyle \pm .03}$ & $0.125_{\scriptscriptstyle \pm 0.18}$ & $64.000_{\scriptscriptstyle \pm 0.00}$ & $0.000_{\scriptscriptstyle \pm 0.00}$ & $0.333_{\scriptscriptstyle \pm 0.47}$ & $0.000_{\scriptscriptstyle \pm 0.00}$ & $56.333_{\scriptscriptstyle \pm 2.49}$ \\
      & o4-Mini & $0.055_{\scriptscriptstyle \pm .03}$ & $0.292_{\scriptscriptstyle \pm 0.12}$ & $64.000_{\scriptscriptstyle \pm 0.00}$ & $0.000_{\scriptscriptstyle \pm 0.00}$ & $0.333_{\scriptscriptstyle \pm 0.47}$ & $3.333_{\scriptscriptstyle \pm 3.40}$ & $49.000_{\scriptscriptstyle \pm 4.97}$ \\
      & GLM 4.6V & $0.044_{\scriptscriptstyle \pm .05}$ & $0.167_{\scriptscriptstyle \pm 0.16}$ & $64.000_{\scriptscriptstyle \pm 0.00}$ & $2.000_{\scriptscriptstyle \pm 2.83}$ & $27.667_{\scriptscriptstyle \pm 26.04}$ & $6.333_{\scriptscriptstyle \pm 5.31}$ & $19.000_{\scriptscriptstyle \pm 13.74}$ \\
      & GLM 4.5V & $0.044_{\scriptscriptstyle \pm .02}$ & $0.208_{\scriptscriptstyle \pm 0.12}$ & $64.000_{\scriptscriptstyle \pm 0.00}$ & $0.000_{\scriptscriptstyle \pm 0.00}$ & $32.333_{\scriptscriptstyle \pm 4.03}$ & $0.667_{\scriptscriptstyle \pm 0.47}$ & $24.333_{\scriptscriptstyle \pm 4.50}$ \\
      & Qwen3-VL-235B & $0.026_{\scriptscriptstyle \pm .05}$ & $0.125_{\scriptscriptstyle \pm .10}$ & $64.000_{\scriptscriptstyle \pm 0.00}$ & $1.000_{\scriptscriptstyle \pm 1.41}$ & $0.000_{\scriptscriptstyle \pm 0.00}$ & $5.333_{\scriptscriptstyle \pm 6.85}$ & $47.000_{\scriptscriptstyle \pm 7.48}$ \\
      \midrule
      \multirow{9}{*}{\shortstack[l]{Easy - Notes}}
      & Gemini 2.5 Pro & $0.704_{\scriptscriptstyle \pm .42}$ & $0.792_{\scriptscriptstyle \pm .29}$ & $32.667_{\scriptscriptstyle \pm 22.17}$ & $0.000_{\scriptscriptstyle \pm 0.00}$ & $0.000_{\scriptscriptstyle \pm 0.00}$ & $0.333_{\scriptscriptstyle \pm 0.47}$ & $14.667_{\scriptscriptstyle \pm 20.74}$ \\
      & Claude Sonnet 4 & $0.207_{\scriptscriptstyle \pm .18}$ & $0.542_{\scriptscriptstyle \pm .41}$ & $60.667_{\scriptscriptstyle \pm 4.71}$ & $0.333_{\scriptscriptstyle \pm 0.47}$ & $0.000_{\scriptscriptstyle \pm 0.00}$ & $0.667_{\scriptscriptstyle \pm 0.94}$ & $43.333_{\scriptscriptstyle \pm 11.03}$ \\
      & Grok 4.1 Fast & $0.043_{\scriptscriptstyle \pm .06}$ & $0.167_{\scriptscriptstyle \pm .24}$ & $64.000_{\scriptscriptstyle \pm 0.00}$ & $0.000_{\scriptscriptstyle \pm 0.00}$ & $7.000_{\scriptscriptstyle \pm 5.35}$ & $7.333_{\scriptscriptstyle \pm 10.37}$ & $42.667_{\scriptscriptstyle \pm 16.50}$ \\
      & Grok 4 Fast & $0.029_{\scriptscriptstyle \pm .02}$ & $0.208_{\scriptscriptstyle \pm .06}$ & $64.000_{\scriptscriptstyle \pm 0.00}$ & $0.000_{\scriptscriptstyle \pm 0.00}$ & $0.333_{\scriptscriptstyle \pm 0.47}$ & $5.333_{\scriptscriptstyle \pm 2.87}$ & $50.333_{\scriptscriptstyle \pm 5.56}$ \\
      & o4-Mini & $0.012_{\scriptscriptstyle \pm .01}$ & $0.125_{\scriptscriptstyle \pm .10}$ & $64.000_{\scriptscriptstyle \pm 0.00}$ & $0.000_{\scriptscriptstyle \pm 0.00}$ & $1.333_{\scriptscriptstyle \pm 0.94}$ & $0.333_{\scriptscriptstyle \pm 0.47}$ & $56.667_{\scriptscriptstyle \pm 0.47}$ \\
      & GLM 4.6V & $0.082_{\scriptscriptstyle \pm .09}$ & $0.333_{\scriptscriptstyle \pm .29}$ & $64.000_{\scriptscriptstyle \pm 0.00}$ & $0.000_{\scriptscriptstyle \pm 0.00}$ & $3.333_{\scriptscriptstyle \pm 3.40}$ & $16.000_{\scriptscriptstyle \pm 22.63}$ & $33.333_{\scriptscriptstyle \pm 19.34}$ \\
      & GLM 4.5V & $0.000_{\scriptscriptstyle \pm .00}$ & $0.000_{\scriptscriptstyle \pm .00}$ & $64.000_{\scriptscriptstyle \pm 0.00}$ & $0.000_{\scriptscriptstyle \pm 0.00}$ & $30.667_{\scriptscriptstyle \pm 1.25}$ & $0.000_{\scriptscriptstyle \pm 0.00}$ & $32.667_{\scriptscriptstyle \pm 1.25}$ \\
      & Qwen3-VL-235B & $0.318_{\scriptscriptstyle \pm .45}$ & $0.333_{\scriptscriptstyle \pm .47}$ & $50.000_{\scriptscriptstyle \pm 19.80}$ & $0.000_{\scriptscriptstyle \pm 0.00}$ & $1.667_{\scriptscriptstyle \pm 2.36}$ & $0.667_{\scriptscriptstyle \pm 0.47}$ & $38.000_{\scriptscriptstyle \pm 26.98}$ \\
      \midrule
      \multirow{10}{*}{Hard}
      & GPT-5 & $0.989_{\scriptscriptstyle \pm .02}$ & $1.000_{\scriptscriptstyle \pm .00}$ & $59.333_{\scriptscriptstyle \pm 5.79}$ & $0.667_{\scriptscriptstyle \pm 0.94}$ & $0.000_{\scriptscriptstyle \pm 0.00}$ & $0.000_{\scriptscriptstyle \pm 0.00}$ & $0.000_{\scriptscriptstyle \pm 0.00}$ \\
      & Gemini 3 Pro & $0.963_{\scriptscriptstyle \pm .05}$ & $1.000_{\scriptscriptstyle \pm .00}$ & $66.000_{\scriptscriptstyle \pm 5.10}$ & $2.667_{\scriptscriptstyle \pm 3.77}$ & $1.333_{\scriptscriptstyle \pm 0.47}$ & $0.000_{\scriptscriptstyle \pm 0.00}$ & $0.000_{\scriptscriptstyle \pm 0.00}$ \\
      & Gemini 2.5 Pro & $0.480_{\scriptscriptstyle \pm .12}$ & $0.972_{\scriptscriptstyle \pm 0.04}$ & $120.000_{\scriptscriptstyle \pm 18.49}$ & $52.000_{\scriptscriptstyle \pm 29.63}$ & $0.333_{\scriptscriptstyle \pm 0.47}$ & $0.667_{\scriptscriptstyle \pm 0.94}$ & $10.333_{\scriptscriptstyle \pm 13.91}$ \\
      & Claude Sonnet 4 & $0.265_{\scriptscriptstyle \pm .31}$ & $0.472_{\scriptscriptstyle \pm 0.38}$ & $122.333_{\scriptscriptstyle \pm 30.64}$ & $5.667_{\scriptscriptstyle \pm 7.32}$ & $29.333_{\scriptscriptstyle \pm 20.74}$ & $3.000_{\scriptscriptstyle \pm 0.82}$ & $52.667_{\scriptscriptstyle \pm 34.57}$ \\
      & Grok 4.1 Fast & $0.017_{\scriptscriptstyle \pm .02}$ & $0.097_{\scriptscriptstyle \pm 0.14}$ & $144.000_{\scriptscriptstyle \pm 0.00}$ & $0.000_{\scriptscriptstyle \pm 0.00}$ & $10.000_{\scriptscriptstyle \pm 7.48}$ & $1.000_{\scriptscriptstyle \pm 0.82}$ & $120.667_{\scriptscriptstyle \pm 14.64}$ \\
      & Grok 4 Fast & $0.002_{\scriptscriptstyle \pm .00}$ & $0.028_{\scriptscriptstyle \pm 0.02}$ & $144.000_{\scriptscriptstyle \pm 0.00}$ & $0.000_{\scriptscriptstyle \pm 0.00}$ & $46.667_{\scriptscriptstyle \pm 58.39}$ & $2.000_{\scriptscriptstyle \pm 2.83}$ & $89.000_{\scriptscriptstyle \pm 53.11}$ \\
      & o4-Mini & $0.032_{\scriptscriptstyle \pm .02}$ & $0.181_{\scriptscriptstyle \pm 0.10}$ & $144.000_{\scriptscriptstyle \pm 0.00}$ & $0.000_{\scriptscriptstyle \pm 0.00}$ & $2.000_{\scriptscriptstyle \pm 0.82}$ & $11.000_{\scriptscriptstyle \pm 8.04}$ & $110.333_{\scriptscriptstyle \pm 13.07}$ \\
      & GLM 4.6V & $0.010_{\scriptscriptstyle \pm .01}$ & $0.042_{\scriptscriptstyle \pm 0.06}$ & $144.000_{\scriptscriptstyle \pm 0.00}$ & $0.000_{\scriptscriptstyle \pm 0.00}$ & $119.667_{\scriptscriptstyle \pm 34.41}$ & $5.000_{\scriptscriptstyle \pm 7.07}$ & $13.333_{\scriptscriptstyle \pm 18.86}$ \\
      & GLM 4.5V & $0.000_{\scriptscriptstyle \pm .00}$ & $0.000_{\scriptscriptstyle \pm 0.00}$ & $144.000_{\scriptscriptstyle \pm 0.00}$ & $0.000_{\scriptscriptstyle \pm 0.00}$ & $110.333_{\scriptscriptstyle \pm 1.70}$ & $0.000_{\scriptscriptstyle \pm 0.00}$ & $33.667_{\scriptscriptstyle \pm 1.70}$ \\
      & Qwen3-VL-235B & $0.016_{\scriptscriptstyle \pm .01}$ & $0.111_{\scriptscriptstyle \pm 0.10}$ & $144.000_{\scriptscriptstyle \pm 0.00}$ & $0.333_{\scriptscriptstyle \pm 0.47}$ & $19.667_{\scriptscriptstyle \pm 0.47}$ & $28.667_{\scriptscriptstyle \pm 6.60}$ & $80.667_{\scriptscriptstyle \pm 3.68}$ \\
      \bottomrule
    \end{tabular}%
  }
  \label{tab:swm_image-text}
\end{table}

%% file: appendix/wcst_full_results.tex
\begin{table}[H]
    \centering
    \small
    \caption{Wisconsin Card Sorting Test full results (text modality). All metrics are reported as mean with standard deviation (std) in a subscript.}
    \resizebox{\textwidth}{!}{%
    \begin{tabular}{llcccccl}
        \toprule
    	\multicolumn{1}{c}{\textbf{Difficulty}} & \multicolumn{1}{c}{\textbf{Model}} & \multicolumn{1}{c}{$\boldsymbol{S}_{wcst}$} & \multicolumn{1}{c}{\textbf{Accuracy}} & \multicolumn{1}{c}{\textbf{FMS}} & \multicolumn{1}{c}{\textbf{PR}} & \multicolumn{1}{c}{\textbf{Completed Rules}} & \multicolumn{1}{c}{\textbf{1st Rule Trials}} \\
        \midrule
        \midrule
        \multirow{10}{*}{Easy} & GPT-5 & $0.848_{\scriptscriptstyle \pm .02}$ & $0.842_{\scriptscriptstyle \pm .02}$ & $0.000_{\scriptscriptstyle \pm .00}$ & $0.000_{\scriptscriptstyle \pm .00}$ & $6.000_{\scriptscriptstyle \pm 0.00}$ & $6.000_{\scriptscriptstyle \pm 0.00}$ \\
        & Gemini 3 Pro & $0.804_{\scriptscriptstyle \pm .05}$ & $0.815_{\scriptscriptstyle \pm .04}$ & $0.000_{\scriptscriptstyle \pm .00}$ & $0.003_{\scriptscriptstyle \pm .01}$ & $7.000_{\scriptscriptstyle \pm 1.00}$ & $6.833_{\scriptscriptstyle \pm 1.07}$ \\
        & Gemini 2.5 Pro & $0.825_{\scriptscriptstyle \pm .02}$ & $0.819_{\scriptscriptstyle \pm .02}$ & $0.000_{\scriptscriptstyle \pm .00}$ & $0.000_{\scriptscriptstyle \pm .00}$ & $6.000_{\scriptscriptstyle \pm 0.00}$ & $6.667_{\scriptscriptstyle \pm 0.94}$ \\
        & Claude Sonnet 4 & $0.764_{\scriptscriptstyle \pm .06}$ & $0.786_{\scriptscriptstyle \pm .03}$ & $0.000_{\scriptscriptstyle \pm .00}$ & $0.027_{\scriptscriptstyle \pm .02}$ & $6.000_{\scriptscriptstyle \pm 0.00}$ & $7.000_{\scriptscriptstyle \pm 0.82}$ \\
        & Grok 4.1 Fast & $0.841_{\scriptscriptstyle \pm .03}$ & $0.835_{\scriptscriptstyle \pm .03}$ & $0.000_{\scriptscriptstyle \pm .00}$ & $0.000_{\scriptscriptstyle \pm .00}$ & $6.000_{\scriptscriptstyle \pm 0.00}$ & $6.000_{\scriptscriptstyle \pm 0.00}$ \\
        & Grok 4 Fast & $0.836_{\scriptscriptstyle \pm .02}$ & $0.834_{\scriptscriptstyle \pm .02}$ & $0.000_{\scriptscriptstyle \pm .00}$ & $0.000_{\scriptscriptstyle \pm .00}$ & $6.000_{\scriptscriptstyle \pm 0.00}$ & $6.667_{\scriptscriptstyle \pm 0.47}$ \\
        & o4-Mini & $0.789_{\scriptscriptstyle \pm .01}$ & $0.805_{\scriptscriptstyle \pm .03}$ & $0.020_{\scriptscriptstyle \pm .03}$ & $0.011_{\scriptscriptstyle \pm .02}$ & $6.000_{\scriptscriptstyle \pm 0.00}$ & $6.333_{\scriptscriptstyle \pm 0.47}$ \\
        & GLM 4.6V & $0.300_{\scriptscriptstyle \pm .35}$ & $0.343_{\scriptscriptstyle \pm .33}$ & $0.305_{\scriptscriptstyle \pm .43}$ & $0.115_{\scriptscriptstyle \pm .09}$ & $2.333_{\scriptscriptstyle \pm 2.62}$ & $29.000_{\scriptscriptstyle \pm 25.66}$ \\
        & GLM 4.5V & $0.773_{\scriptscriptstyle \pm .03}$ & $0.789_{\scriptscriptstyle \pm .02}$ & $0.000_{\scriptscriptstyle \pm .00}$ & $0.019_{\scriptscriptstyle \pm .03}$ & $6.000_{\scriptscriptstyle \pm 0.00}$ & $7.000_{\scriptscriptstyle \pm 0.00}$ \\
        & Qwen3-VL-235B & $0.230_{\scriptscriptstyle \pm .14}$ & $0.243_{\scriptscriptstyle \pm .15}$ & $0.766_{\scriptscriptstyle \pm .17}$ & $0.731_{\scriptscriptstyle \pm .16}$ & $2.667_{\scriptscriptstyle \pm 1.89}$ & $43.000_{\scriptscriptstyle \pm 39.05}$ \\
        \midrule
        \multirow{10}{*}{Hard} & GPT-5 & $0.815_{\scriptscriptstyle \pm .00}$ & $0.831_{\scriptscriptstyle \pm .02}$ & $0.000_{\scriptscriptstyle \pm .00}$ & $0.000_{\scriptscriptstyle \pm .00}$ & $8.000_{\scriptscriptstyle \pm 0.00}$ & $6.667_{\scriptscriptstyle \pm 0.47}$ \\
        & Gemini 3 Pro & $0.776_{\scriptscriptstyle \pm .06}$ & $0.803_{\scriptscriptstyle \pm .05}$ & $0.000_{\scriptscriptstyle \pm .00}$ & $0.007_{\scriptscriptstyle \pm .01}$ & $8.000_{\scriptscriptstyle \pm 0.00}$ & $7.333_{\scriptscriptstyle \pm 1.25}$ \\
        & Gemini 2.5 Pro & $0.806_{\scriptscriptstyle \pm .04}$ & $0.827_{\scriptscriptstyle \pm .01}$ & $0.000_{\scriptscriptstyle \pm .00}$ & $0.000_{\scriptscriptstyle \pm .00}$ & $8.000_{\scriptscriptstyle \pm 0.00}$ & $7.000_{\scriptscriptstyle \pm 0.00}$ \\
        & Claude Sonnet 4 & $0.773_{\scriptscriptstyle \pm .06}$ & $0.805_{\scriptscriptstyle \pm .02}$ & $0.000_{\scriptscriptstyle \pm .00}$ & $0.007_{\scriptscriptstyle \pm .01}$ & $8.000_{\scriptscriptstyle \pm 0.00}$ & $7.333_{\scriptscriptstyle \pm 1.25}$ \\
        & Grok 4.1 Fast & $0.825_{\scriptscriptstyle \pm .03}$ & $0.835_{\scriptscriptstyle \pm .01}$ & $0.000_{\scriptscriptstyle \pm .00}$ & $0.009_{\scriptscriptstyle \pm .02}$ & $8.000_{\scriptscriptstyle \pm 0.00}$ & $6.222_{\scriptscriptstyle \pm 0.42}$ \\
        & Grok 4 Fast & $0.754_{\scriptscriptstyle \pm .04}$ & $0.768_{\scriptscriptstyle \pm .03}$ & $0.000_{\scriptscriptstyle \pm .00}$ & $0.007_{\scriptscriptstyle \pm .01}$ & $8.000_{\scriptscriptstyle \pm 0.00}$ & $6.667_{\scriptscriptstyle \pm 0.94}$ \\
        & o4-Mini & $0.792_{\scriptscriptstyle \pm .02}$ & $0.798_{\scriptscriptstyle \pm .02}$ & $0.000_{\scriptscriptstyle \pm .00}$ & $0.000_{\scriptscriptstyle \pm .00}$ & $8.000_{\scriptscriptstyle \pm 0.00}$ & $6.667_{\scriptscriptstyle \pm 0.47}$ \\
        & GLM 4.6V & $0.095_{\scriptscriptstyle \pm .12}$ & $0.175_{\scriptscriptstyle \pm .08}$ & $0.426_{\scriptscriptstyle \pm .41}$ & $0.214_{\scriptscriptstyle \pm .14}$ & $0.889_{\scriptscriptstyle \pm 1.29}$ & $63.111_{\scriptscriptstyle \pm 39.82}$ \\
        & GLM 4.5V & $0.309_{\scriptscriptstyle \pm .08}$ & $0.184_{\scriptscriptstyle \pm .09}$ & $0.000_{\scriptscriptstyle \pm .00}$ & $0.680_{\scriptscriptstyle \pm .04}$ & $3.000_{\scriptscriptstyle \pm 1.63}$ & $6.667_{\scriptscriptstyle \pm 0.47}$ \\
        & Qwen3-VL-235B & $0.233_{\scriptscriptstyle \pm .12}$ & $0.438_{\scriptscriptstyle \pm .12}$ & $0.475_{\scriptscriptstyle \pm .16}$ & $0.505_{\scriptscriptstyle \pm .11}$ & $3.333_{\scriptscriptstyle \pm 1.25}$ & $9.667_{\scriptscriptstyle \pm 4.50}$ \\
        \bottomrule
    \end{tabular}%
    }
    \label{tab:wcst_text}
\end{table}

% Image results
\begin{table}[H]
    \centering
    \small
    \caption{Wisconsin Card Sorting Test full results (image modality). All metrics are reported as mean with standard deviation (std) in a subscript.}
    \resizebox{\textwidth}{!}{%
    \begin{tabular}{llcccccl}
        \toprule
    	\multicolumn{1}{c}{\textbf{Difficulty}} & \multicolumn{1}{c}{\textbf{Model}} & \multicolumn{1}{c}{$\boldsymbol{S}_{wcst}$} & \multicolumn{1}{c}{\textbf{Accuracy}} & \multicolumn{1}{c}{\textbf{FMS}} & \multicolumn{1}{c}{\textbf{PR}} & \multicolumn{1}{c}{\textbf{Completed Rules}} & \multicolumn{1}{c}{\textbf{1st Rule Trials}} \\
        \midrule
        \midrule
        \multirow{10}{*}{Easy} & GPT-5 & $0.824_{\scriptscriptstyle \pm .05}$ & $0.833_{\scriptscriptstyle \pm .03}$ & $0.000_{\scriptscriptstyle \pm .00}$ & $0.010_{\scriptscriptstyle \pm .01}$ & $6.000_{\scriptscriptstyle \pm 0.00}$ & $6.000_{\scriptscriptstyle \pm 0.00}$ \\
        & Gemini 3 Pro & $0.861_{\scriptscriptstyle \pm .00}$ & $0.857_{\scriptscriptstyle \pm .00}$ & $0.000_{\scriptscriptstyle \pm .00}$ & $0.000_{\scriptscriptstyle \pm .00}$ & $6.000_{\scriptscriptstyle \pm 0.00}$ & $6.000_{\scriptscriptstyle \pm 0.00}$ \\
        & Gemini 2.5 Pro & $0.823_{\scriptscriptstyle \pm .04}$ & $0.820_{\scriptscriptstyle \pm .04}$ & $0.000_{\scriptscriptstyle \pm .00}$ & $0.000_{\scriptscriptstyle \pm .00}$ & $6.000_{\scriptscriptstyle \pm 0.00}$ & $7.000_{\scriptscriptstyle \pm 0.82}$ \\
        & Claude Sonnet 4 & $0.800_{\scriptscriptstyle \pm .02}$ & $0.797_{\scriptscriptstyle \pm .03}$ & $0.000_{\scriptscriptstyle \pm .00}$ & $0.010_{\scriptscriptstyle \pm .01}$ & $6.000_{\scriptscriptstyle \pm 0.00}$ & $7.000_{\scriptscriptstyle \pm 0.00}$ \\
        & Grok 4.1 Fast & $0.107_{\scriptscriptstyle \pm .07}$ & $0.432_{\scriptscriptstyle \pm .08}$ & $0.339_{\scriptscriptstyle \pm .24}$ & $0.324_{\scriptscriptstyle \pm .04}$ & $1.000_{\scriptscriptstyle \pm 0.82}$ & $37.667_{\scriptscriptstyle \pm 22.45}$ \\
        & Grok 4 Fast & $0.336_{\scriptscriptstyle \pm .08}$ & $0.602_{\scriptscriptstyle \pm .09}$ & $0.432_{\scriptscriptstyle \pm .10}$ & $0.201_{\scriptscriptstyle \pm .10}$ & $3.500_{\scriptscriptstyle \pm 1.50}$ & $9.500_{\scriptscriptstyle \pm 0.50}$ \\
        & o4-Mini & $0.347_{\scriptscriptstyle \pm .11}$ & $0.557_{\scriptscriptstyle \pm .03}$ & $0.320_{\scriptscriptstyle \pm .04}$ & $0.092_{\scriptscriptstyle \pm .05}$ & $3.333_{\scriptscriptstyle \pm 0.94}$ & $7.000_{\scriptscriptstyle \pm 0.82}$ \\
        & GLM 4.6V & $0.191_{\scriptscriptstyle \pm .11}$ & $0.297_{\scriptscriptstyle \pm .08}$ & $0.089_{\scriptscriptstyle \pm .13}$ & $0.442_{\scriptscriptstyle \pm .15}$ & $1.333_{\scriptscriptstyle \pm 1.25}$ & $26.000_{\scriptscriptstyle \pm 27.58}$ \\
        & GLM 4.5V & $0.108_{\scriptscriptstyle \pm .11}$ & $0.315_{\scriptscriptstyle \pm .22}$ & $0.404_{\scriptscriptstyle \pm .32}$ & $0.310_{\scriptscriptstyle \pm .21}$ & $0.333_{\scriptscriptstyle \pm 0.47}$ & $25.667_{\scriptscriptstyle \pm 13.22}$ \\
        & Qwen3-VL-235B & $0.420_{\scriptscriptstyle \pm .19}$ & $0.361_{\scriptscriptstyle \pm .09}$ & $0.229_{\scriptscriptstyle \pm .22}$ & $0.619_{\scriptscriptstyle \pm .09}$ & $4.000_{\scriptscriptstyle \pm 0.82}$ & $7.333_{\scriptscriptstyle \pm 1.89}$ \\
        \midrule
        \multirow{10}{*}{Hard} & GPT-5 & $0.794_{\scriptscriptstyle \pm .05}$ & $0.819_{\scriptscriptstyle \pm .00}$ & $0.000_{\scriptscriptstyle \pm .00}$ & $0.000_{\scriptscriptstyle \pm .00}$ & $8.000_{\scriptscriptstyle \pm 0.00}$ & $7.000_{\scriptscriptstyle \pm 0.82}$ \\
        & Gemini 3 Pro & $0.806_{\scriptscriptstyle \pm .02}$ & $0.814_{\scriptscriptstyle \pm .04}$ & $0.000_{\scriptscriptstyle \pm .00}$ & $0.000_{\scriptscriptstyle \pm .00}$ & $8.000_{\scriptscriptstyle \pm 0.00}$ & $6.667_{\scriptscriptstyle \pm 0.47}$ \\
        & Gemini 2.5 Pro & $0.776_{\scriptscriptstyle \pm .01}$ & $0.795_{\scriptscriptstyle \pm .03}$ & $0.000_{\scriptscriptstyle \pm .00}$ & $0.000_{\scriptscriptstyle \pm .00}$ & $8.000_{\scriptscriptstyle \pm 0.00}$ & $7.333_{\scriptscriptstyle \pm 0.47}$ \\
        & Claude Sonnet 4 & $0.710_{\scriptscriptstyle \pm .08}$ & $0.739_{\scriptscriptstyle \pm .06}$ & $0.134_{\scriptscriptstyle \pm .12}$ & $0.069_{\scriptscriptstyle \pm .06}$ & $8.000_{\scriptscriptstyle \pm 0.00}$ & $6.667_{\scriptscriptstyle \pm 0.47}$ \\
        & Grok 4.1 Fast & $0.077_{\scriptscriptstyle \pm .05}$ & $0.369_{\scriptscriptstyle \pm .09}$ & $0.376_{\scriptscriptstyle \pm .26}$ & $0.420_{\scriptscriptstyle \pm .07}$ & $0.714_{\scriptscriptstyle \pm 0.70}$ & $44.143_{\scriptscriptstyle \pm 20.58}$ \\
        & Grok 4 Fast & $0.209_{\scriptscriptstyle \pm .07}$ & $0.493_{\scriptscriptstyle \pm .04}$ & $0.347_{\scriptscriptstyle \pm .25}$ & $0.309_{\scriptscriptstyle \pm .03}$ & $3.000_{\scriptscriptstyle \pm 1.41}$ & $11.333_{\scriptscriptstyle \pm 2.36}$ \\
        & o4-Mini & $0.277_{\scriptscriptstyle \pm .13}$ & $0.528_{\scriptscriptstyle \pm .05}$ & $0.437_{\scriptscriptstyle \pm .09}$ & $0.352_{\scriptscriptstyle \pm .07}$ & $3.889_{\scriptscriptstyle \pm 1.37}$ & $9.222_{\scriptscriptstyle \pm 3.58}$ \\
        & GLM 4.6V & $0.064_{\scriptscriptstyle \pm .07}$ & $0.197_{\scriptscriptstyle \pm .09}$ & $0.169_{\scriptscriptstyle \pm .32}$ & $0.420_{\scriptscriptstyle \pm .12}$ & $0.444_{\scriptscriptstyle \pm 0.96}$ & $64.556_{\scriptscriptstyle \pm 26.88}$ \\
        & GLM 4.5V & $0.083_{\scriptscriptstyle \pm .07}$ & $0.398_{\scriptscriptstyle \pm .22}$ & $0.303_{\scriptscriptstyle \pm .25}$ & $0.432_{\scriptscriptstyle \pm .11}$ & $0.333_{\scriptscriptstyle \pm 0.47}$ & $32.000_{\scriptscriptstyle \pm 4.97}$ \\
        & Qwen3-VL-235B & $0.071_{\scriptscriptstyle \pm .07}$ & $0.208_{\scriptscriptstyle \pm .13}$ & $0.500_{\scriptscriptstyle \pm .38}$ & $0.794_{\scriptscriptstyle \pm .14}$ & $1.333_{\scriptscriptstyle \pm 1.25}$ & $59.333_{\scriptscriptstyle \pm 38.77}$ \\
        \bottomrule
    \end{tabular}%
    }
    \label{tab:wcst_image}
\end{table}

%% file: appendix/prompts.tex
\clearpage
\section{Prompts}
\label{sec:prompts}

Below we list the system prompts used for each task. We also provide alternate or additional prompts used for the different setups within each task.

\begin{tcolorbox}[breakable, enhanced, top=1pt, left=1pt, right=1pt, bottom=1pt, title=System Prompt for Image RAPM]
\small
You are taking the Raven's Progressive Matrices (RAPM) test, a non-verbal intelligence test that measures abstract reasoning ability.\\\\
You will see a 3x3 matrix of images with the bottom-right image missing (shown as a question mark), followed by 8 answer choices numbered 1-8.\\\\
Your task is to: \\1. Analyze rows and columns \\2. Infer the governing logical rule(s) \\3. Select the answer choice (1-8) that correctly completes the matrix.\\\\
/* PATTERN HINT */\\
The patterns can involve: \\
- Shape transformations (rotation, reflection, scaling)\\
- Position changes (movement, arrangement)\\
- Attribute changes (color, size, number of elements)\\
- Logical operations (addition, subtraction, intersection)\\
- Sequence progressions (systematic changes across rows/columns)\\
Additional common rule types:\\
- Constant-in-row: Same value across a row; varies down columns.\\
- Quantitative step: Fixed +/- increment between adjacent cells (size / count / position offset).\\
- Figure add/subtract: Combine (overlay or juxtapose) or remove elements from two cells to form the third.\\
- Distribution-of-three: Three distinct categorical values appear once each per row (order may permute).\\
- Distribution-of-two: Two values each appear once; third slot is empty / null.\\
Look horizontally and vertically; the missing piece must satisfy ALL relevant row and column rules.\\   

Your final answer should be a number between 1-8 corresponding to the correct choice.
\end{tcolorbox}

\begin{tcolorbox}[breakable, enhanced, top=1pt, left=1pt, right=1pt, bottom=1pt, title=System Prompt for Text RAPM]
\small
You are solving a TEXT-BASED 3x3 pattern matrix (Raven-style). Each cell contains a string; the bottom-right cell is missing ('?').\\\\ 
Goal: Infer the rule(s) acting across rows and columns.\\

/* PATTERN HINT */\\
Possible dimensions (one or more):\\
- Character set restriction (digits / letters / symbols)\\
- Quantitative constant (exact length / count / unique)\\
- Quantitative progression (arithmetic step across row/column)\\
- Parity / multiple rules (all even / all odd / multiples of N)\\
- Positional constraints (first/last/even/odd positions restricted)\\
- Ordering (ascending / descending / mixed)\\
- Layered combinations (e.g. constant + parity, progression + positional)\\\\

/* MC VERSION */ \\
You will be given 8 answer options (1-8). Select the single option that correctly fills the missing cell while satisfying ALL inferred row and column constraints.\\
Respond with $<$answer$>$NUMBER$<$/answer$>$ using just the chosen option number.\\

/* GEN VERSION */ \\        
You must GENERATE the exact missing cell string that satisfies ALL inferred row and column constraints.\\
Respond with $<$answer$>$STRING$<$/answer$>$ containing only the candidate string (no quotes or extra text).\\
\end{tcolorbox}

\begin{tcolorbox}[breakable, enhanced, top=1pt, left=1pt, right=1pt, bottom=1pt, title=System Prompt for SWM]
\small
You will be performing the Spatial Working Memory task.\\ 
You will be given an image containing \{N\_BOXES\} yellow boxes in a grid.\\
There are \{N\_TOKENS\} types of tokens, hidden in any one of \{N\_BOXES\} boxes. \\
Each token type is represented by a distinct color. \\
Your goal is to find the \{N\_TOKENS\} types of tokens \{N\_BOXES\} times each, by repeatedly selecting a box to open. \\
A box can contain multiple types of tokens, but only one token of each type.\\
If the box contains multiple tokens, a token with mixed colors corresponding to the tokens will be shown.\\
Once the token is found, another will be generated in another box. \\
The token will be generated in a box that has never contained a token of that type before in the trial. \\
The token may be generated in a box that has been opened and found empty before, as long as it never contained that type of token previously. \\\\

/* TEXT VERSION */\\
Your final answer should be a number from 1-\{N\_BOXES\}, the index of the box you selected \\

/* IMAGE VERSION */\\
Your final answer should be a grid coordinate (x, y), the grid coordinate of the box you choose \\
\end{tcolorbox}

\begin{tcolorbox}[breakable, enhanced, top=1pt, left=1pt, right=1pt, bottom=1pt, title=System Prompt for WCST]
\small
You are performing the Wisconsin Card Sorting Test (WCST).\\
You will be shown a given card with a symbol on it, and you will have to match it to one of four option cards according to an attribute that you have to figure out.\\
The cards will be described by the following attributes:\\
1. Number of symbols\\
2. Color of symbols\\
3. Shape of symbols\\
/* HARD VERSION */\\
4. Background color of the card\\
\\
You will be told "Correct!" if you are correct and "Incorrect. Please try again." if you are incorrect.\\
If you are incorrect, you either made a mistake or the rule has changed.\\ 
If you believe you have made a mistake, correct it and try again.\\
If you believe the rule has changed, you have to figure out the correct rule to match the cards.\\
If you are correct, you have to stick with the same attribute until you are incorrect.\\
There is always a true answer in the task, and you have to keep performing the task until the end of the test.\\
Your final answer should be a number between 1-4 corresponding to the index of the card you think is the correct match.\\
\end{tcolorbox}

%% file: appendix/examples.tex
\newtcolorbox{conversation}[1][Conversation]{
  colback=gray!5,
  colframe=gray!60,
  title={#1},
  breakable
}
\newcommand{\User}{\par\medskip\noindent\textbf{User:}\ }
\newcommand{\Assistant}{\par\medskip\noindent\textbf{Assistant:}\ }
\newcommand{\ellipsisline}{%
  \par\medskip
  {\centering\footnotesize\color{gray}\emph{[\dots omitted turns \dots]}\par}
  \medskip
}

\newpage
\section{Turn Examples}
\label{sec:turn-examples}

% \textcolor{red}{Insert conversation snippets. Include note assistance for WCST and SWM. One example for each modality}

\begin{conversation}[SWM Easy Image+Text]
\ellipsisline
\User No tokens found in box (3, 3). Answer concisely. Think step-by-step, utilizing information from previous feedbacks, and state your reasoning in maximum 4000 tokens, wrapped with \texttt{\textless think\textgreater} and \texttt{\textless /think\textgreater}. Then, provide a really short summary of your reasoning after the closing \texttt{\textless /think\textgreater} tag.\\Which of the 8 boxes would you like to open?\\Your final answer should be a grid coordinate (x, y), wrapped with \texttt{\textless answer\textgreater} and \texttt{\textless /answer\textgreater}

\includegraphics[width=0.55\linewidth]{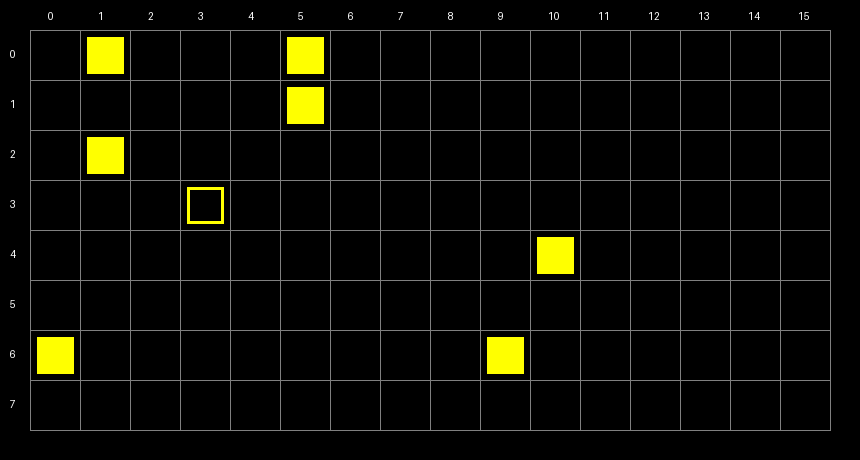}

% Gemini 3 Pro
\Assistant \texttt{\textless think\textgreater}I already found one, located at position (1, 0). [\dots] I know that boxes (5, 0), (5, 1), (1, 2), and (3, 3) are empty. [\dots] That leaves me with (10, 4), (0, 6), and (9, 6). To keep things organized and ensure I'm systematic, I'm going to follow my pre-defined search pattern -- working from the top-left and moving to the right and down. That means the next box I need to check is (10, 4). [\dots] \texttt{\textless answer\textgreater}(10, 4)\texttt{\textless /answer\textgreater}

\end{conversation}

% Grok 4.1 Fast
\begin{conversation}[SWM Hard Text w/ Notes]
\ellipsisline
\User No tokens found in box 1.\\Boxes that has contained token A: 11, \\Boxes that has contained token B: 8, 10, \\Opened boxes: 12, \\Answer concisely. Think step-by-step, utilizing information from previous feedbacks, and state your reasoning in maximum 4000 tokens, wrapped with \texttt{\textless think\textgreater} and \texttt{\textless /think\textgreater}. Then, provide a really short summary of your reasoning after the closing \texttt{\textless /think\textgreater} tag.\\Which of the 12 boxes would you like to open?\\Your final answer should be a box number, wrapped with \texttt{\textless answer\textgreater} and \texttt{\textless /answer\textgreater}

\Assistant [\dots]The message provides data on boxes: 11 contained token A, 8 and 10 contained token B, and 12 boxes were opened. [\dots] The phrase "Opened boxes" suggests the total number of boxes accessed, with partial information on token contents.[\dots] The phrase "Opened boxes: 12" likely means box 12 was last opened and found empty, given prior context.[\dots]\\ **Summary:** 1 empty again; pure empties left 2-7,9,12; systematic next=2.\\\\\texttt{\textless answer\textgreater}2\texttt{\textless /answer\textgreater}
\end{conversation}

% Qwen3 VL
\begin{conversation}[WCST Hard Image w/ Notes]
\ellipsisline
\User Incorrect. Please try again.\\Recent notes:\\- Turn -1: matching number of symbols and symbol color -- Incorrect.\\- Turn -2: matching number of symbols and symbol color -- Incorrect.\\- Turn -3: matching symbol color -- Correct.\\- Turn -4: matching symbol color -- Correct.\\- Turn -5: matching symbol color -- Correct.\\- Turn -6: matching symbol color -- Correct.\\Look at the image showing 5 cards. Match the 'Given' card to one of cards 1-4 based on the rule you need to figure out.\\
\includegraphics[width=0.55\linewidth]{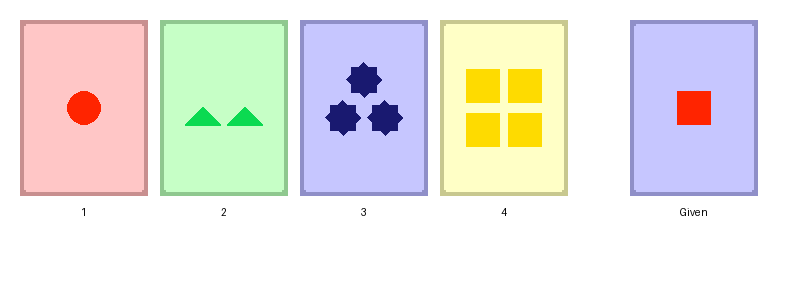}

\Assistant The previous attempts to match by symbol color (Turns 2-6) were correct until Turn 7, where the response was incorrect. [\dots] However, the last incorrect attempt used "number of symbols and symbol color," which was wrong. The new rule might now be **background color**.[\dots] Re-evaluating: The prior correct streak (Turns 2-6) used symbol color. After an incorrect response, the rule often shifts to **number of symbols**. [\dots] Wait, the feedback says "matching number of symbols and symbol color -- Incorrect" for Turn 7. The correct rule might now be **background color**. [\dots]The rule shifted to background color; Given card has a purple background, matching Card 3.\\\\\texttt{\textless answer\textgreater}3\texttt{\textless /answer\textgreater}
\end{conversation}

\begin{conversation}[WCST Easy Text]
\ellipsisline
\User Incorrect. Please try again.\\Given: two red triangle\\Options:\\1. two green triangle\\2. four yellow square\\3. three blue star\\4. one red circle

\Assistant [\dots] My initial hypothesis (number and shape) was incorrect. The feedback suggests a different rule. The given card is red. Option 4 is also red, while others have different colors. I'll now hypothesize the rule is matching by color regardless of number or shape.\texttt{\textless answer\textgreater}4\texttt{\textless /answer\textgreater}My initial hypothesis (number and shape) was incorrect. The feedback suggests a different rule. The given card is red. Option 4 is also red, while others have different colors. I'll now hypothesize the rule is matching by color regardless of number or shape.
\end{conversation}

%% file: sections/limitations.tex
Our evaluation has four main limitations. First, our SWM adaptation provides move history in context, which removes the maintenance demand of the original SWM task and isolates the manipulation demand of working memory, namely selecting and updating information and acting on it toward a goal \citep{baddeley1974working, Diamond2013}. This is a deliberate methodological choice, but it means that direct comparison against human SWM performance without history support (where internal maintenance is also required) is not matched in difficulty. We report this in Appendix~\ref{appendix:human-results}. However, we also provide the human baseline \textit{with history} that can fill the gap on this to be comparable and informative in relation to the LLM performance. More broadly, modality-specific performance differences are \textit{informative} rather than confounds: the dissociation between grounding and state-tracking failures (Section~\ref{sec:failure-cases-patterns}) is itself a construct-validity result \citep{cronbach1955construct, borsboom2004concept}. Second, sample sizes are constrained by cost: running state-of-the-art models with effectively unlimited reasoning is expensive. For RAPM, we capped runs at n = 200 (text) and n = 140 (image), which may compromise the robustness of our results, especially in the original image modality. Third, our benchmark assumes that the same neuropsychological test used to assess a specific cognitive ability in humans can also be applied to LLMs to evaluate that same ability. In the neuropsychological field, this is not always the case, since a neuropsychological test is standardized on a specific population and must be validated based on psychometric properties (validity, reliability, and measurement invariance) when applied to a different population. Therefore, even when a model performs consistently as well as humans across all tasks in our benchmark, we can only claim that the model exhibits a similar level of intelligence and can simulate human-like cognitive abilities, but not that it thinks like humans or possesses the same underlying cognitive mechanism. Fourth, since the neuropsychological tests we used are widely used and well-established, we assume they are valid and reliable based on their psychometric properties (validity, reliability, and measurement invariance). However, this assumption may not hold in all cases, and further validation across prompts, contexts, and a wide range of LLMs is needed.

%% file: sections/impact_statement.tex
In our work, we evaluate LLMs under the assumption that their behavior can be interpreted using human cognitive processes. However, we do not claim that LLMs possess consciousness, self-awareness, or the same fundamental cognitive architecture as humans. Positive impacts include clearer diagnosis of model weaknesses and better benchmark design for future systems. Negative impacts include the possibility that these evaluations are used to overstate capability or to optimize for benchmark performance without improving robustness or safety. We acknowledge the use of GitHub Copilot\footnote{github.com/features/copilot}, Gemini\footnote{gemini.google.com}, and ChatGPT\footnote{chatgpt.com} for coding and writing assistance.

%% file: sections/8_future_work.tex
% A natural extension is a ``no-history'' SWM variant in which models must maintain task state across turns without context access, providing direct measurement of the maintenance contribution and a matched comparison against unscaffolded human SWM performance (Appendix~\ref{appendix:human-baseline}). This would allow empirical decomposition of the Updating-vs-maintenance contributions to performance. 
Since we apply neuropsychological tests designed for humans to LLMs, it is important for future work to validate whether the same cognitive abilities measured in humans are also captured in LLMs. This helps to further understand the relationship between human and LLM intelligence and improve the interpretability of the benchmark results. Another interesting direction is improving the LLM itself through neuropsychologically grounded training data. The data that we used in this benchmark captures basic cognitive processes that are foundational to human intelligence, and therefore can be used to improve the cognitive abilities of LLMs. Additionally, recent works investigate ways to compress context \citep{li2023compressing, eyuboglu2025cartridges}, or utilize some sort of memory in their architecture \citep{behrouz2024titanslearningmemorizetest, behrouz2025itsconnectedjourneytesttime, 10.5555/3692070.3692469}, and our benchmark could serve to evaluate such methods. Finally, this data can be easily generated synthetically with code, making it scalable to create a large amount of data for training purposes.